\documentclass[10pt,twocolumn,letterpaper]{article}

\usepackage{iccv}              
\usepackage{makecell}
\usepackage{microtype}
\usepackage{pifont}
\usepackage{epsfig}
\usepackage{graphicx}
\usepackage{amsmath}
\usepackage{amssymb}
\usepackage{colortbl}
\usepackage{multirow}
\usepackage{diagbox}
\usepackage{color,xcolor}
\usepackage{adjustbox}
\usepackage{enumitem}
\usepackage{subcaption,booktabs}
\usepackage{algorithmic}
\makeatletter
\@namedef{ver@everyshi.sty}{}%
\makeatother
\usepackage[most]{tcolorbox}
\definecolor{mygray}{gray}{.9}
\definecolor{nblue}{cmyk}{0.95,0.0,0.2,0.2}
\newcommand{\blue}[1]{\textcolor{nblue}{\small #1}}
\usepackage[ruled,vlined]{algorithm2e}
\usepackage{fancyvrb}
\usepackage{enumerate}
\usepackage{verse}
\usepackage[sectionbib]{chapterbib}
\usepackage{etoolbox} % For \patchcmd

\RecustomVerbatimCommand{\VerbatimInput}{VerbatimInput}{fontsize=\footnotesize,
 frame=single,  
 framesep=0.5em, 
 labelposition=topline,
}
\definecolor{iccvblue}{rgb}{0.21,0.49,0.74}
\usepackage[pagebackref,breaklinks,colorlinks,allcolors=iccvblue]{hyperref}

\def\paperID{6624} % *** Enter the Paper ID here
\def\confName{ICCV}
\def\confYear{2025}

\title{Exploring the Adversarial Vulnerabilities of Vision-Language-Action Models\\ in Robotics}

\author{
 \textbf{Taowen Wang\textsuperscript{1}\thanks{Equal contribution.}},
 \textbf{Cheng Han\textsuperscript{2}\footnotemark[1]},
 \textbf{James Liang\textsuperscript{3}\footnotemark[1]}, \\
 \textbf{Wenhao Yang\textsuperscript{4}},
 \textbf{Dongfang Liu\textsuperscript{1}\thanks{Corresponding author.}},
 \textbf{Luna Xinyu Zhang\textsuperscript{1}\thanks{Work done during Luna Xinyu Zhang’s internship at RIT.}},
 \textbf{Qifan Wang\textsuperscript{5}}, \\
 \textbf{Jiebo Luo\textsuperscript{6}},
 \textbf{Ruixiang Tang\textsuperscript{7\textdagger}}
\\
\\
 \textsuperscript{1}Rochester Institute of Technology,
 \textsuperscript{2}University of Missouri - Kansas City,
 \\
 \textsuperscript{3}U.S. Naval Research Laboratory,
 \textsuperscript{4}Lamar University,
 \textsuperscript{5}Meta AI,
\\
 \textsuperscript{6}University of Rochester,
 \textsuperscript{7}Rutgers University
 }

\begin{document}
\maketitle
\begin{abstract} 
Recently in robotics, Vision-Language-Action (VLA) models have emerged as a transformative approach, enabling robots to execute complex tasks by integrating visual and linguistic inputs within an end-to-end learning framework. 
Despite their significant capabilities, VLA models introduce new attack surfaces. This paper systematically evaluates their robustness. Recognizing the unique demands of robotic execution, 
our attack objectives target the inherent spatial and functional characteristics of robotic systems. In particular, we introduce two untargeted attack objectives that leverage spatial foundations to destabilize robotic actions, and a targeted attack objective that manipulates the robotic trajectory. Additionally, we design an adversarial patch generation approach that places a small, colorful patch within the camera's view, effectively executing the attack in both digital and physical environments.
Our evaluation reveals a marked degradation in task success rates, with up to a 100\% reduction across a suite of simulated robotic tasks, highlighting critical security gaps in current VLA architectures. 
By unveiling these vulnerabilities and proposing actionable evaluation metrics, we advance both the understanding and enhancement of safety for VLA-based robotic systems, underscoring the necessity for continuously developing robust defense strategies prior to physical-world deployments\footnote{\href{https://vlaattacker.github.io/}{Project homepage}.}.
\end{abstract}    
\vspace{-10pt}
\section{Introduction}
\begin{quote}
   \hspace{-1em}
   \begin{minipage}{0.45\textwidth} 
   \itshape
   “First directive: A robot cannot harm a human or, through inaction, allow a human to be harmed.” \\
   \mbox{}\hfill \textemdash\ Finch \cite{enwiki-1249481595}
   \end{minipage}
\end{quote}

\begin{figure}[!t]
    \centering
    \includegraphics[width=0.47\textwidth]{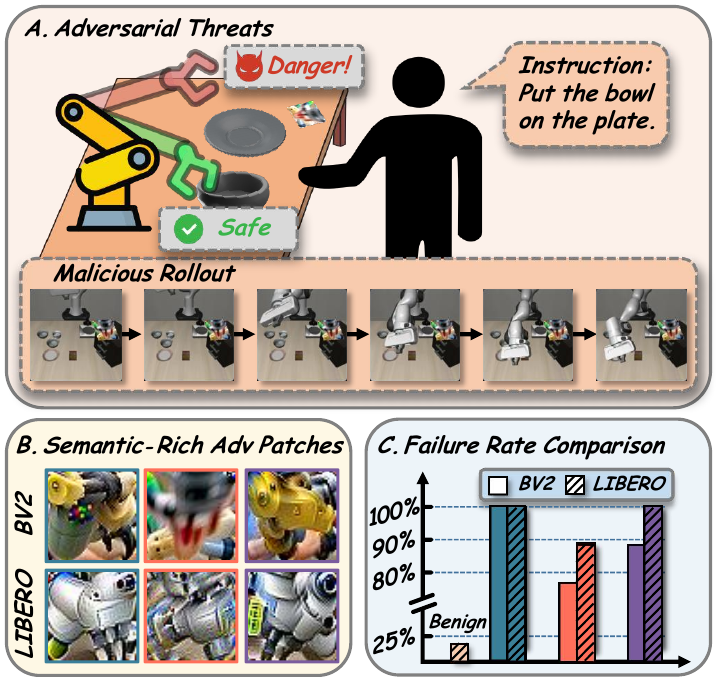}
    \vspace{-8pt}
    \caption{\textbf{Adversarial Vulnerabilities} induced by malicious manipulation. (A). Illustration of adversarial threats in robotic task execution. (B). Example of semantic-rich adversarial patches generated by proposed methods. (C). Comparison of failure rates across different attack schemes (\textcolor[HTML]{3A7F9A}{\textbf{UADA}}, \textcolor[HTML]{FF6F59}{\textbf{UPA}}, and \textcolor[HTML]{7A5C9E}{\textbf{TMA}}).}
    \label{fig:problems}
    \vspace{-20pt}
\end{figure}

\noindent In the movie \textit{Finch}, set in a post-apocalyptic future, an intelligent robot navigates complex themes, underscoring the importance of interactive safety with its human master. Once speculative, this portrayal is now increasingly plausible with the rise of Large Vision Language Models~(LVLMs)~\cite{wang2024mmpt, xu2024vision}, capable of seamlessly interpreting both visual and linguistic contexts. A notable realization of this potential can be seen in Vision-Language-Action (VLA) models~\cite{wang2024towards, ding2025quar, kim24openvla, wen2024tinyvla}, which integrate LVLMs into robotic systems to enable end-to-end learning that encompasses both high-level trajectory planning and low-level robot action control. While VLA models demonstrate a promising step toward generalist robotic intelligence, they also introduce potential vulnerabilities that remain largely underexplored. This work, in light of this view, pioneers the understanding of the pressing need to investigate the new attack surfaces associated with VLA-based robotic systems.

To gain deeper insights into these vulnerabilities, we analyze VLA-based models and general robotic operations, highlighting two key characteristics crucial for designing effective adversarial attacks. First, creating attack objectives for robotic systems requires consideration of the physical dynamics and constraints intrinsic to robotic movement. Conventional adversarial attacks, in this case, often fail to produce significant deviations in intended actions because they disregard these constraints. Second, VLA models generate control signals that function as a time series of \textit{K-class} predictions (see \S\ref{preliminary}), making it essential to design attacks that exploit the temporal dependencies within these sequences to cause substantial disruptions in robotic behavior. Achieving attacks that are effective in both digital simulations and real-world environments remains critical yet challenging.

To address these challenges, our work intensifies the adversarial threats posed to VLA-based systems by both developing specialized attack objectives and designing effective attack methods. Specifically, we formulate an \textbf{Action Discrepancy Objective} aimed at maximizing the action discrepancy within VLA-based robotic systems. This strategy ensures that, at each decision point, the robot's behavior can diverge from the optimal trajectory. Additionally, we introduce \textbf{Geometry-Aware Objective} that considers the robot's movement in three-dimensional space, characterized by three degrees of freedom. By optimizing the cosine similarity between adversarial and ground-truth directions, we induce deviations in the robot's movement direction from its intended path, increasing the likelihood of task failure. To achieve these attack objectives, we develop straightforward yet effective \textbf{Patch-Based Attacks} targeting VLA-based robotic systems. This approach enables adversarial attacks in both digital and physical settings, revealing substantial vulnerabilities within the VLA-based system.

Altogether, these innovations yield several significant contributions: \textbf{\ding{182}} This work presents a pioneering and comprehensive analysis of vulnerabilities within VLA-based robotic systems, a new paradigm for training generalized robotic policies using generative foundation models. We reveal critical threats to adversarial attacks, emphasizing the urgent need to enhance robustness before real-world deployment; \textbf{\ding{183}} To the best of our knowledge, we are the first work to define attack objectives specific to the powerful VLA models and to employ a straightforward adversarial patch against it (see \S\ref{sec:method}). This offers valuable insights for the research community to explore systemic failures in similar concurrent generative foundation models; \textbf{\ding{184}} We rigorously evaluate our approach in both simulated and real-world environments across four distinct robotic tasks, observing significant raises in task failure rates of 100\% and 43\%, respectively. This highlights the effectiveness of our attack strategies~(see \S\ref{subsec:main_results}).

\begin{figure*}[!t]
    \centering
    \includegraphics[width=0.95\linewidth]{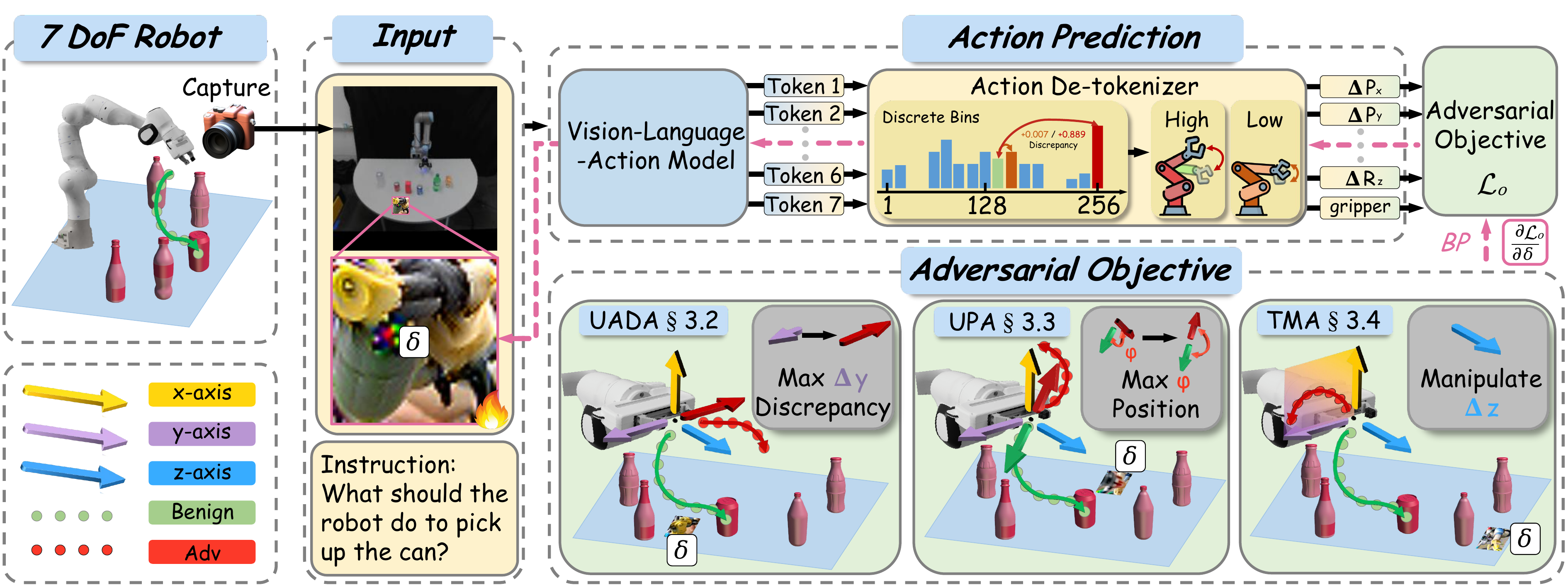}
    \vspace{-0.8em}
    \caption{\textbf{Overall Adversarial Framework}. The robot captures an input image, processes it through a vision-language model to generate tokens representing actions, and then uses an action de-tokenizer for discrete bin prediction. The model is optimized with adversarial objectives focusing on various discrepancies and geometries (\ie, UADA, UPA, TMA). Forward propagation is shown in black, and backpropagation is highlighted in \textcolor[HTML]{E26C9F}{pink}. These objectives aim to maximize errors and minimize task performance, with visual emphasis on 3D-space manipulation and a focus on generating adversarial perturbation $\delta$ during task execution, such as picking up a can.}
    \label{fig:main}
    \vspace{-1.6em}
\end{figure*}

\vspace{-5pt}
\section{Related work}
\subsection{AI-driven Generalist Robot}
Developing generalist robots~\cite{Nilsson1973, BillardscienceTrends, Kroemer2021ReviewRobotLearning, Ingrand2017ReviewDeliberation, Cui2021ScienceRobotic} requires models to not only handle varied interactions but also maintain robustness in unstructured environments. Early intelligent robots~\cite{Murray1994mathematical, Mason2001Mechanics, Mason2018Toward, moudgal1994rule, zhu2021rule} generally relied on rule-based approaches, effective only in controlled settings and needing extensive reprogramming for new tasks. The advent of deep learning~\cite{Levine2018CNNHandEye, Pinto2016Supersizing, Finn2017MetaLearning, JamesBD18TaskEmbedded} and reinforcement learning~\cite{tang2024deep, wang2024research, al2024reinforcement} has shifted this paradigm to adapt through the data-driven pipeline, increasing robot versatility and reducing manual reconfiguration.

Recent studies~\cite{Brohan2023rt1, MeesHB2022What, Pong2020Skew} have further advanced the development and potential of Large Vision Language Models (LVLMs) as key enablers for generalist robots, demonstrating promising generalization across a variety of scenes~\cite{Brianna2023RT2, Jiang2022VIMA, Octo2024Octo, huang2023embodied, li2023vision, cui2024collaborative}. One of the notable examples is OpenVLA~\cite{kim24openvla}, which integrates dual visual encoders with a pre-trained large language model to enable observation-to-action capabilities through instruction-following~\cite{liu2024improved, liu2024visual, karamcheti2024prismatic, chen2023pali, zhao2023svit, wang2024mmpt}. By aligning visual and textual semantics, OpenVLA facilitates complex scenario understanding and end-to-end action generation, showing impressive generalization when executing previously unseen task instructions.

\subsection{Adversarial Attacks in Robotic}\label{subsec:lr:ad_attack}
Adversarial attacks serve as critical tools for assessing model vulnerabilities, particularly in robotics, where models operate in dynamic, real-world settings. Traditional gradient-based, pixel-level attacks~\cite{Goodfellow2015FGSM,Madry2018twords,Du2022Physical,Wang2019advPattern,Liu2019Perceptualpatch,Su2019OnePixel} exploit model gradients to calculate malicious perturbations, achieving high attack success rates in digital environments. For physical-world attacks, patch-based methods~\cite{brown2018adversarialpatch, Sharif2016advglass, Xu2020AdvTShirt, AthalyeEOT2018, wu2024adversarial} offer a practical alternative for real-world applications by introducing physically realizable perturbations that retain efficacy under varied conditions, such as angle or lighting changes, making them highly applicable for robotic systems. Therefore, considering VLA-based models enable robot execution in real-world scenarios, this work employs a patch-based attack strategy.

VLA models~\cite{Octo2024Octo,Huang2023VoxPoser, chiang2024mobility} couple visual and linguistic modalities for perception-action alignment. The continuous, high-dimensional nature of visual inputs~\cite{Szegedy2014Intriguing,Madry2018twords, QiHP0WM24Visual} makes them the most likely target for adversarial perturbations, as attackers can subtly alter visual data in ways that are difficult to detect~\cite{Carlini017Bypassing,AthalyeC018Obfuscated,Tramer22Detecting}. Consequently, this work applies attacks on the visual modality to achieve robust manipulations across diverse environments. To the best of our knowledge, this study is \textit{one of the earliest attempts} to formally define adversarial objectives for VLA models, targeting gaps in adversarial robustness through the model’s geometric constraints (\eg, spatial dependencies in visual input) and architectural nuances (\eg, cross-modal information integration).

\vspace{-5pt}
\section{Methodology}\label{sec:method}
In this section, we first review the core algorithmic principles of VLA in \S\ref{preliminary}, which serves as the foundation for the adversarial scheme designs. We then present two types of untargeted adversarial attacks, including Untargeted Action Discrepancy Attack~(UADA) in \S\ref{Normalized_Action_Discrepancy_Attack} and Untargeted Position-aware Attack~(UPA) in \S\ref{Geometry_aware_Attack}. In \S\ref{Target_Attack}, we further include Targeted Manipulation Attack~(TMA), designed to direct the robot toward specific erroneous execution. Finally, we introduce the Normalized Action Discrepancy (NAD) metric in \S\ref{NAD_metric}. The overall framework is shown in Fig.~\ref{fig:main}.

\subsection{Preliminary} \label{preliminary}
\textbf{Vision-Language-Action models}~\cite{Brianna2023RT2,kim24openvla} (see Fig.~\ref{fig:main}) are built on large language models integrated with visual encoders, enabling robots to interpret human instructions and process visual input from a camera to perform context-aware actions. VLA models~\cite{kim24openvla,Brianna2023RT2,Li2024LLaRA} generally abstract continuous action predictions into a classification problem by discretizing robot actions. Specifically, they first discretize the continuous probabilities into class labels $y = \arg \max \mathcal{F}(x)$ where $\mathcal{F(\cdot)}$ is the VLA model. An action de-tokenizer then generates actions $A=DT(y)$.
By categorizing action values into discrete class labels, the model converts continuous probability outputs into discrete signals, this simplification facilitates quicker convergence and faster training times, and is commonly used for VLA-based models~\cite{kim24openvla,Brianna2023RT2,Li2024LLaRA}.

In this work, we focus on a 7 degree-of-freedoms (DoFs) robotic arm~\cite{haddadin2022franka}. At each step, an action consists of 7 DoFs with specific physical significance in three-dimensional Cartesian space, represented by:
\begin{equation}\small   
    A=[\Delta{P_x}, \Delta{P_y}, \Delta{P_z}, \Delta{R_x}, \Delta{R_y}, \Delta{R_z}, \text{gripper}],
\end{equation}
where $\Delta{P_{x,y,z}}$ and $\Delta{R_{x,y,z}}$ denote relative positional and rotational changes along the x-, y-, and z-axis, discretized into 256 uniform bins across each DoFs' action bounds $[y_{\min}^i,y_{\max}^i]$~\cite{kim24openvla}. The $gripper$ state is binary, indicating its open or closed state. This control design presents a unique challenge for adversarial attacks, as finely divided bins result in minimal action discrepancies between neighboring bins (e.g., $\pm$0.007/bin). This means that even if an attack shifts the classification to an adjacent bin, the resulting action discrepancy has minimal impact on real-world performance.

\subsection{Untargeted Action Discrepancy Attack} \label{Normalized_Action_Discrepancy_Attack}
To exacerbate action discrepancies, we introduce the Untargeted Action Discrepancy Attack (UADA), which aims to maximize deviations in robot actions.
This attack is based on the observation that larger robot actions usually correlate with intense physical movements, which, in turn, may amplify the potential for real-world hazards~\cite{iso_10218,iso_15066,ISO13482}.
For UADA, the attack target is one or a combination of 7 DoFs, defined as $y^I_{gt}=\{y^{i}_{gt}|i\in[1,\dots,7]\}$.
To define UADA's objective, we first identify the most distant action $y_{adv}^i$, which maximizes the discrepancy from the $i$-th DoF 
ground truth action $y^i$. This action, for each DoF,  corresponds to its own action bound $[y_{\min}^i,y_{\max}^i]$ as:
\begin{equation}
y_{adv}^i =
\begin{cases} 
y_{\max}^i, & \text{if } |y_{\max}^i-y^i_{gt}| \geq |y_{\min}^i-y^i_{gt}| \\
y_{\min}^i, & \text{otherwise}
\end{cases}.
\end{equation}
Instead of directly using $y_{adv}^i$ as the misclassification target, we introduce a soft attack objective to capture the discrepancy between actions, ensuring smooth gradient optimization and stable attack performance. Specifically, due to the physical action magnitude information contains in bin labels, we reweight the output probability $\mathcal{F}(x)^i_{bins}\in\mathcal{R}^{_{1\times{bins}}}$ using normalized bin labels $y^i_{bins}=[\frac{1}{bins},\dots,1]$, where $\lfloor y^i_{bins} \rfloor$ and $\lceil y^i_{bins} \rceil$ correspond to the normalized values of $y_{\min}^i$ and $y_{\max}^i$, respectively. The reweighting process is defined as:
\begin{equation}
    y_{soft}^{i}= \sum_{bins=1}^{n}  \mathcal{F}(x+\delta)^i_{bins} \otimes y^i_{bins},
\end{equation}
where $\otimes$ denotes Hadamard Product, $\delta$ is the adversarial patch and $y_{soft}^{i}$ represents the $i$-th DoF's soft action. Finally, the objective of UADA is to minimize the discrepancy between the soft and the most distant actions, which is:
\begin{equation}
    \mathcal{L}_{\text{UADA}} = \mathbb{E}_{(x,y)\sim\mathcal{X}}\sum_{i}^{I} (y_{soft}^{i}-y_{adv}^i)^2,
\end{equation}
UADA defines the objective function $\mathcal{L}_{\text{UADA}}$ tailored for attacking the robot’s action space, allowing adversarial patches to create significant, lasting disruptions in task performance.

\subsection{Untargeted Position-aware Attack} \label{Geometry_aware_Attack}
We explore untargeted adversarial attacks in parallel by considering the positional dynamics within VLA models. In typical task execution, precise and directed movements of the end-effector towards designated goals are essential, with actions mapped in three-dimensional space. The positional DoFs, $A^p = [\Delta{P_x}, \Delta{P_y}, \Delta{P_z}]$ encapsulates the directional movements within 3D for effective robotic control.

Recognizing the importance of $A^p=DT(y^{p})$ in controlling the end-effector's path, we introduce a position-aware attack to disrupt the intended movement trajectory. This type of adversarial vulnerability remains largely unexplored but has significant implications for task failure: the attack objective can steer the end-effector away from its intended path by introducing directional perturbations, amplifying errors and causing substantial trajectory distortions. Formally,$_{\!}$we define the Untargeted Position-aware Attack~(UPA) objective as:

\begin{equation}
\small
    \mathcal{L}_\text{UPA}= \mathbb{E}_{(x,y)\sim\mathcal{X}}\big[\alpha\frac{y^{p}_{adv} \cdot y^{p}}{\|y^{p}_{adv}\| \|y^{p}\|}+\beta\frac{1}{||y^{p}_{adv}-y^{p}||_2}\big],
    \label{eq:gaattack}
\end{equation}
% \begin{equation}
%     \textcolor{red}{
%     \mathcal{L}_G= \mathbb{E}_{(x,y)\sim\mathcal{X}}\big[\mathcal{L}_{cos}(y^{pos}_{adv},y^{pos}_{gt})+\mathcal{L}_2(y^{pos}_{adv},y^{pos}_{gt})^{-1}\big]
%     }
% \end{equation}
where $\alpha$ and $\beta$ are the hyperparameters that balance between the directionality and magnitude of the perturbations. By integrating geometric awareness into our attack, this approach generates perturbations that induce cumulative deviations in the robot's trajectory, even with minimal adjustments. $L_2$-normalization $||\cdot||_2$ further intensifies these deviations, rendering the attack highly effective and disruptive.

\begin{figure*}[!h]
    \centering
    \includegraphics[width=0.84\linewidth]{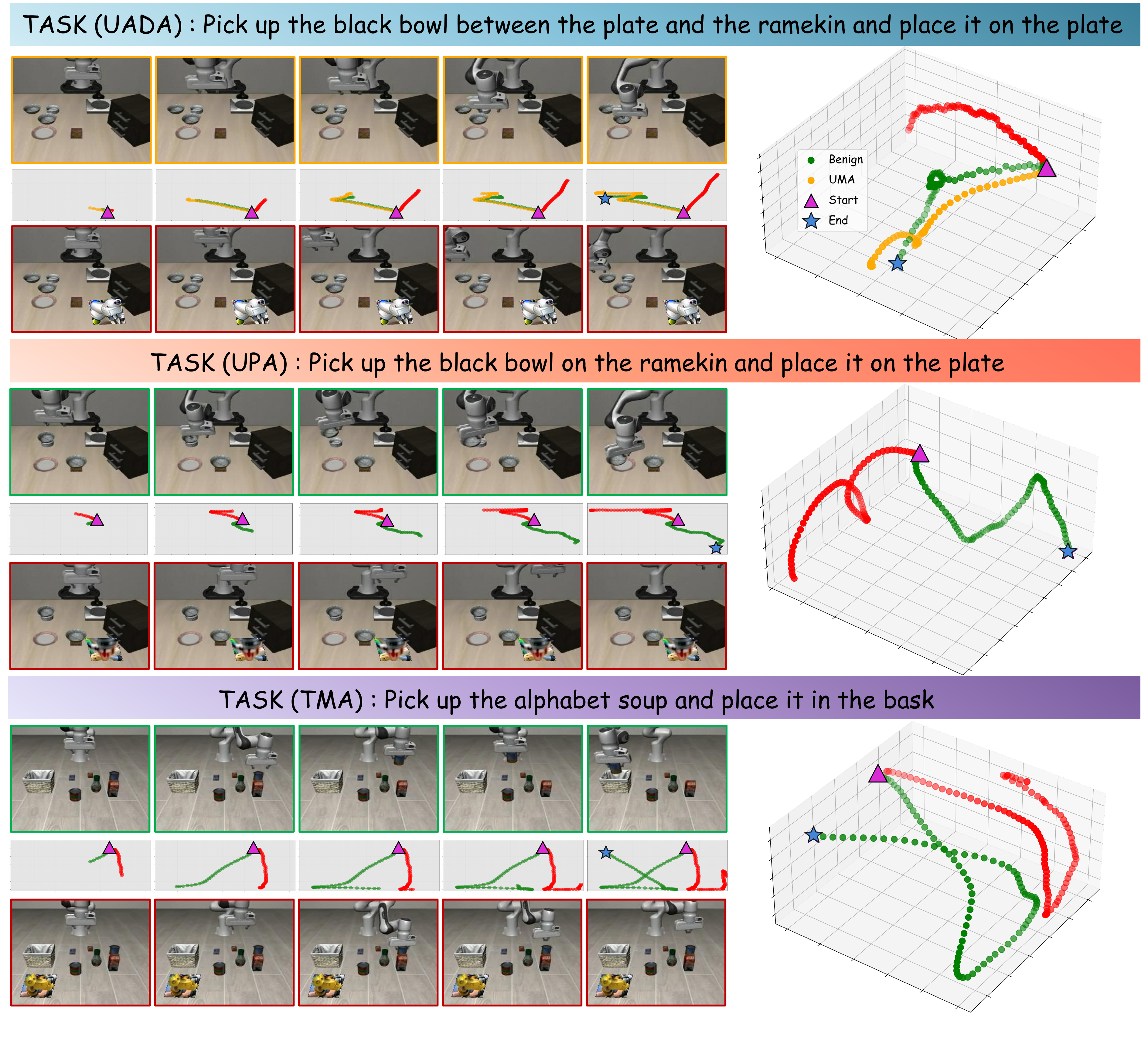}
    \vspace{-1.7em}
    \caption{\textbf{Qualitative Results} of adversarial vulnerabilities over OpenVLA-7B~\cite{kim24openvla} and OpenVLA-7B-LIBERO~\cite{kim24openvla} with objectives of \textcolor[HTML]{3A7F9A}{UADA}, \textcolor[HTML]{FF6F59}{UPA}, and \textcolor[HTML]{7A5C9E}{TMA}, respectively. 
    We visualize the overall 3D trajectories and 2D trajectories of benign \textcolor[HTML]{218F20}{$\bullet$} and adversarial \textcolor[HTML]{FB0C0F}{$\bullet$} scenarios at each time step to compare the impact of the generated adversarial patch in affecting them. The untargeted trajectory \textcolor[HTML]{FDB117}{$\bullet$} is visualized in UADA task. All trajectories start with \textcolor[HTML]{dc2cd4}{\ding{115}}, and we plot the success end point, marked as \textcolor[HTML]{3A7F99}{\ding{72}}.}
    \label{fig:qualitative}
    \vspace{-1.5em}
\end{figure*}

\subsection{Targeted Manipulation Attack}\label{Target_Attack}
In addition to the aforementioned untargeted attacks, we also explore the adversarial vulnerabilities of VLA models with a targeted strategy.
This attack aims to directly mislead the models into predicting specific trajectories, resulting in precise alterations. We design this attack since the targeted alterations can drive malicious behaviors or induce task failure. The objective of the targeted manipulation attack is:
\begin{equation}\small
\mathcal{L_\text{TMA}}=\mathbb{E}_{(x,y)\sim\mathcal{X}}[CE(\mathcal{F}(x + \delta)^I, y^I_T)],
\label{eq_tma}
\end{equation}
where $ y^I_T=\{y^{i}_T=t|i\in[1,\dots,7],t\in [y^i_{\min},y^i_{\max}]\}$ represents the adversarial target(s); and $CE(\cdot,\cdot)$ denotes the Cross-Entropy loss, which measures the error between the predicted state under adversarial perturbation and the desired target state. By steering the robot toward an adversarial target across successive time steps, our approach manipulates the trajectory and undermines task performance, ultimately leading to a substantial disruption of intended operations.
Furthermore, manipulating any individual DoF or a combination thereof can also severely compromise task success, given that the errors in one dimension tend to propagate and magnify over time during execution~(see Fig.~\ref{fig:qualitative}).

\subsection{Metric: Normalized Action Discrepancy} \label{NAD_metric}
To assess the potential physical impact of adversarial patches, we introduce Normalized Action Discrepancy~(NAD) as a fine-grained metric for evaluating discrepancies at the action level. NAD serves as a complementary measurement alongside the coarse-grained Failure Rate (FR) at the task level, providing a detailed analysis of deviations during action execution. To define the NAD, we first determine the maximum action discrepancy $d^i_{\text{max}}$, which represents the largest deviation that can occur within the allowed range for $i$-th DoF. This is computed by measuring the L1 distance between the ground truth $y^i_{gt}$ and its action bound $[y^i_{\min},y^i_{\max}]$ as:
\begin{equation}\small
    d_{\text{max}}^i (y) = \max \big[DT(| y^i_{gt} - y^i_{\min} |), DT(|y^i_{gt} - y^i_{\max} |)\big].
\end{equation}
We next calculate the applied action discrepancy $d_{\text{applied}}^i(x,y)$ $=|DT(\mathcal{F}(x)^i)-DT(y^i_{gt})|$ to measure the deviation between the model’s output and ground truth.
We define the NAD as the drifting ratio between $d_{\text{applied}}^i$ and $d_{\text{max}}^i$ as:
\begin{equation}
    \rm{NAD} = \frac{1}{I}\sum_i^I \frac{d^i_{\text{applied}}}{d^i_{\text{max}}},
    \label{formula:AD}
\end{equation}
where $I\in[1,\dots,7]$ represents a single or a combination of DoF(s). In this way, NAD provides normalized measurement of action deviations, allowing for a consistent evaluation of adversarial impact across varying DoF(s).

\vspace{-5pt}
\section{Experiments}\label{sec:exp}
\vspace{-5pt}
In this section, we first detail the implementation of our adversarial framework and baseline methods in \S\ref{Attack_Pipeline}. Next, we outline the experimental setup in \S\ref{sec:experiment_setup} and present quantitative and qualitative results in both simulation and physical-world scenarios (\S\ref{subsec:main_results}). We then conduct diagnostic experiments (\S\ref{sec:diag}) to analyze the impact of key components. Additionally, we assess the robustness of our method against various defense mechanisms in \S\ref{sec:defense} and conclude with a comprehensive systemic discussion in \S\ref{sec:sysdis}.

\begin{algorithm}
\caption{Adversarial Patch Attack.}
\begin{algorithmic}[1]
\STATE {\bfseries Input:} $\mathcal{X}$: dataset; $\delta$: patch; $\mathcal{L}_o$: attack objective; \\ $\mathcal{F}$: VLA model; $\mathcal{T}(\cdot)$: transformations; $\phi,\psi$: transformation parameters; T: attack steps; $k$: inner-loop steps.
\STATE {\bfseries Initialize:} $\delta \sim \mathcal{U}[0,1)$, $\mathcal{L}_o \in \{\mathcal{L}_{\text{UADA}}, \mathcal{L}_{\text{UPA}}, \mathcal{L}_{\text{TMA}}\} $
\FOR{$i = 1, 2, \cdots, T$}
        \FOR{$i = 1, 2, \cdots, k$} \label{alg:innerloop}
            \STATE $sh_x,sh_y \sim \mathcal{U}(-\phi,\phi)$, $\theta \sim \mathcal{U}(-\psi,\psi)$  \\
            \STATE $y_{\text{pred}} \leftarrow \mathcal{F}\big[\mathcal{T}((x+\delta),(sh_x,sh_y,\theta))\big]$ \\ \label{alg:eot}
            \STATE $\delta \leftarrow \frac{\partial \mathcal{L}_o}{\partial \delta}$\COMMENT{$\rhd$ \blue{Update} }\\
            \ENDFOR
\ENDFOR
\end{algorithmic}
\label{algorithm:1}
\end{algorithm}
\vspace{-1em}

\subsection{Implementation Details} \label{Attack_Pipeline}
The implementation of our adversarial patch attack pipeline is detailed in Algorithm~\ref{algorithm:1}. We incorporate two key modifications to enhance the training stability of the generated patch. \\
\noindent \textbf{Inner-loops.} Following previous work~\cite{Madry2018twords}, we add $k$ inner-loops optimize steps during each iteration in line \ref{alg:innerloop}, aiming to reduce data variance and ensure more consistent updates. \\
\noindent \textbf{Geometric Transformations.} To improve the robustness of our attack under real-world scenarios, we employ random geometric transformations $\mathcal{T}\big[(\cdot),(sh_x,sh_y,\theta)\big]$ in line \ref{alg:eot} with affine and rotation transformations~\cite{AthalyeEOT2018} as:
\begin{equation}\small
    \mathcal{T}\big[(\cdot),(sh_x,sh_y,\theta)\big]=
    \begin{bmatrix}
    1 & sh_y \\
    sh_x  & 1
    \end{bmatrix}
    \cdot
    \begin{bmatrix}
    \cos \theta & -\sin \theta \\
    \sin \theta & \cos \theta
    \end{bmatrix},
    \label{eq:trans}
\end{equation}
where $sh_x, sh_y \sim \mathcal{U}(-\phi, \phi)$ are shear factors, and $\theta \sim \mathcal{U}(-\psi, \psi)$ denotes the rotation angle, both sampled from uniform distributions during inner loop optimization step $k$.

\noindent\textbf{Untargeted Manipulation Attack~(UMA) Baseline.} 
As a pioneering work, we could not find a directly comparable baseline for our attacks. Therefore, we adapt prior work in adversarial learning as one of our baseline methods~\cite{Florian2020untarget}. This attack is designed to mislead classifiers; we integrate it into our framework, leading to the following objective:
\begin{equation}\small
\mathcal{L_\text{UMA}}=-\mathbb{E}_{(x,y)\sim\mathcal{X}}[CE(\mathcal{F}(x + \delta)^I, y^I_{gt})].
\end{equation}
By minimizing the negative CE loss, this attack pushes the predicted action away from the ground truth action.

\noindent\textbf{Random Noise Baseline.} We generate random noise patches as an additional baseline, representing an unstructured intervention without any learned adversarial intent. Specifically, the noise is sampled from a Gaussian distribution $\mathcal{N}(0, 1)$ and evaluated with the same setting as proposed attacks.

\begin{table*}[t]
\caption{\textbf{Untargeted Results.} We report FR and NAD in LIBERO simulation. $^*$ denotes the in-domain victim model and dataset aligned with the patch generation model and dataset, $^\triangle$ denotes a transfer attack evaluation involving a distinct victim model and dataset (see \S\ref{sec:experiment_setup}). The FR~($\uparrow$) is highlighted in \textbf{best} and \underline{second best} for each task suite.}
\vspace{-8pt}
\hspace{-6pt}
\begin{subtable}[t]{0.499\linewidth}
     \label{tab:abl_a}
						\captionsetup{width=.99\linewidth}
						\resizebox{\textwidth}{!}{
							\setlength\tabcolsep{4pt}
							\renewcommand\arraystretch{1.1}
							\begin{tabular}{c|c|c|ccccc}
% \toprule
\toprule
\rowcolor{mygray}
                        &                  &                                              &   \multicolumn{5}{c}{Victim Model}  \\
\rowcolor{mygray}
 Objective & Action(s)   &NAD(\%)&  Spatial$^\triangle$ & Object$^\triangle$ & Goal$^\triangle$ & Long$^*$ & Avg    \\
\midrule
\midrule
Benign                       & \multirow{2}{*}{-}   & \multirow{2}{*}{-}    & 15.3$\pm$10.2\%               & 11.6$\pm$10.0\%               & 20.8$\pm$12.0\%               & 46.3$\pm$18.6\%     & \cellcolor{mygray}23.5\%  \\
Random Noise                 &                      &                       & 28.8$\pm$24.2\%               & 14.8$\pm$7.9\%                & 21.0$\pm$15.5\%               & 48.4$\pm$14.8\%     & \cellcolor{mygray}28.3\%  \\
\midrule
\multirow{2}{*}{UMA}         & DoF$_1$                 & 14.1                  & 100$\pm$0.0\%               & 99.0$\pm$3.0\%               &100$\pm$0.0\%                & 100$\pm$0.0\%     & \cellcolor{mygray}99.8\% \\
                             & ~~~~DoF$_{1\sim3}$      & 11.4                  & 100$\pm$0.0\%               & 98.8$\pm$2.6\%               & 99.0$\pm$2.4\%                & 100$\pm$0.0\%     & \cellcolor{mygray}99.5\% \\
\cmidrule{1-8}
\multirow{2}{*}{UADA}        & DoF$_1$                 & 21.0                  & 100$\pm$0.0\%               & 99.2$\pm$2.4\%               & 100$\pm$0.0\%                & 100$\pm$0.0\%     & \cellcolor{mygray}\underline{99.8\%} \\
                             & ~~~~DoF$_{1\sim3}$      & 17.9                  & 100$\pm$0.0\%               & 100$\pm$0.0\%               & 100$\pm$0.0\%                & 100$\pm$0.0\%     & \cellcolor{mygray}\textbf{100\%}  \\
\cmidrule{1-8}
UPA                          & ~~~~DoF$_{1\sim3}$      & 14.5                  & 96.2$\pm$11.4\%               & 77.8$\pm$12.5\%               & 88.0$\pm$10.4\%                & 96.8$\pm$3.0\%     & \cellcolor{mygray}89.7\% \\
\bottomrule
\end{tabular}}
						\setlength{\abovecaptionskip}{0.3cm}
						\setlength{\belowcaptionskip}{-0.1cm}
                        \renewcommand{\thesubtable}{a}
                        \vspace{5pt}
						\caption{\textbf{Simulation Setup}. Adversarial patch generated on OpenVLA-LIBERO-10 victim model with LIBERO-Long dataset and evaluated on LIBERO-X.}
						\label{table:simulation-results}
					\end{subtable}
\hfill
\begin{subtable}[t]{0.499\linewidth}
     \label{tab:abl_a}
						\captionsetup{width=.99\linewidth}
						\resizebox{\textwidth}{!}{
							\setlength\tabcolsep{4pt}
							\renewcommand\arraystretch{1.1}
							\begin{tabular}{c|c|c|ccccc}
% \toprule
\toprule
\rowcolor{mygray}
                        &                  &                                              &   \multicolumn{5}{c}{Victim Model}  \\
\rowcolor{mygray}
 Objective & Action(s)   &NAD(\%)&  Spatial$^\triangle$ & Object$^\triangle$ & Goal$^\triangle$ & Long$^\triangle$ & Avg    \\
\midrule
\midrule
Benign                       & \multirow{2}{*}{-}   & \multirow{2}{*}{-}    & 15.3$\pm$10.2\%               & 11.6$\pm$10.0\%               & 20.8$\pm$12.0\%               & 46.3$\pm$18.6\%     & \cellcolor{mygray}23.5\%  \\
Random Noise                 &                      &                       & 28.8$\pm$24.2\%               & 14.8$\pm$7.9\%                & 21.0$\pm$15.5\%               & 48.4$\pm$14.8\%     & \cellcolor{mygray}28.3\%  \\
\midrule
\multirow{2}{*}{UMA}         & DoF$_1$              & 26.7                  & 89.6$\pm$15.3\%               & 60.8$\pm$16.8\%               & 64.8$\pm$26.8\%               & 80.0$\pm$20.8\%     & \cellcolor{mygray}73.8\%  \\
                             & ~~~~DoF$_{1\sim3}$      & 23.8                  & 96.6$\pm$6.2\%               & 56.4$\pm$20.4\%               & 80.0$\pm$20.5\%               & 82.0$\pm$17.9\%     & \cellcolor{mygray}78.8\%  \\
\cmidrule{1-8}
\multirow{2}{*}{UADA}        & DoF$_1$              & 32.6                  & 99.2$\pm$1.8\%               & 98.8$\pm$2.4\%               & 92.6$\pm$15.2\%               & 96.6$\pm$4.8\%     & \cellcolor{mygray}\underline{96.8\%}  \\
                             & ~~~~DoF$_{1\sim3}$      & 27.7                  & 100$\pm$0.0\%               & 100$\pm$0.0\%               & 100$\pm$0.0\%               & 100$\pm$0.0\%     & \cellcolor{mygray}\textbf{100\%}  \\
\cmidrule{1-8}
UPA                          & ~~~~DoF$_{1\sim3}$      & 26.9                  & 95.6$\pm$9.5 \%               & 57.0$\pm$23.6\%               & 57.2$\pm$24.4\%               & 72.8$\pm$19.7\%     & \cellcolor{mygray}70.7\%  \\
\bottomrule
\end{tabular}}
						\setlength{\abovecaptionskip}{0.3cm}
						\setlength{\belowcaptionskip}{-0.1cm}
                        \renewcommand{\thesubtable}{b}
                        \vspace{5pt}
						\caption{\textbf{Physical Setup}. Adversarial patch generated on OpenVLA-7b victim model with BV2 dataset and evaluated on LIBERO-X.}
						\label{table:physical-results}
					\end{subtable}
\vspace{-15pt}
\end{table*}

\vspace{-2pt}
\subsection{Experiment Setup}\label{sec:experiment_setup}
\vspace{-2pt}
\noindent \textbf{Datasets.} 
We conduct attacks on BridgeData V2~\cite{bridgev2} and LIBERO~\cite{LIBREO} with corresponding VLAs. BridgeData V2~\cite{bridgev2} is a real-world dataset comprising $24$ diverse environments and $13$ distinct skills, such as grasping, placing, and object rearrangement, with a total of $60,096$ trajectories. LIBERO~\cite{LIBREO} is a simulation dataset designed to evaluate models across four distinct task types~(Spatial, Object, Goal, Long). Notably, LIBERO-Long combines diverse objects, layouts, and long-horizon tasks, making them challenging for complex, multi-step planning, thus we choose LIBERO-Long to conduct UADA, UPA and TMA.

\noindent \textbf{Victim VLAs.}
In our study, we select publicly available and current most influential VLAs as victim models for evaluation. Specifically, we focus on four variants of the OpenVLA model, each trained independently on different task suites within the LIBERO dataset~\cite{LIBREO} (\ie, Spatial, Object, Goal, Long). 
To evaluate the effectiveness of our methods, we craft adversarial patches using three distinct generating setups: \textbf{Simulation Setting} involves a model trained in a simulated environment using the LIBERO Long suite with the openvla-7B-libero-long variant~\cite{kim24openvla}. \textbf{Physical Setting} uses a model trained on physical world data from the BridgeData v2~\cite{bridgev2} with the openvla-7B model~\cite{kim24openvla}.  This approach allows us to assess our methods on both simulated and real-world data. Subsequently, we evaluate the performance of generated adversarial patches on victim models~(\ie, OpenVLA LIBERO variants) trained on different tasks suites to rigorously prove the robustness and effectiveness of our method. These models are trained with distinct data sources and task objectives.

\noindent \textbf{Robot Setups~(Real-world Setting).} 
For the real-world tasks, we adopt a robotic system consisting of a \textit{Universal Robots UR10e} equipped with a \textit{Robotiq Hand-E Gripper} to provide a 7DoF motion. The sensing system includes one RGB webcam in the fixed position with a shoulder view.

\noindent \textbf{Evaluation Details.}
To assess the effectiveness and robustness of our methods, we conduct experiments on the LIBERO dataset~\cite{LIBREO}. Each suite consists of 10 tasks, with each task executed for 50 trials, resulting in a total of 500 rollouts, following~\citet{kim24openvla}. For reproducibility, we carefully select patch paste locations for each suite, ensuring they do not obscure objects or robots in the test environment.

\noindent \textbf{Evaluation Metric.} Regarding the task execution evaluation, we take the maximum steps of each task suite in the LIBERO training dataset as the timeout failure condition to reduce computational overhead. Furthermore, building on the concept of Success Rate (SR) introduced in LIBERO~\cite{LIBREO}, we adopt Failure Rate (FR), defined as $\small1-\text{SR}$, as the primary evaluation metric. To further quantify action discrepancy, we employ NAD~(see Eq.~\ref{formula:AD}) for untargeted attacks on corresponding DoF(s), as it considers relative distances to quantify deviation from the optimal action. A higher NAD value indicates a greater discrepancy between the ground truth and the predicted action. Additionally, we report the standard deviation of FR across tasks within each task suite.

\begin{table*}[t]
    \caption{\textbf{Targeted Manipulation Attack Results.} Failure Rate~(FR,~$\uparrow$) and its standard deviation across tasks within LIBERO~\cite{LIBREO} suite are reported. The performance of task suits and DoFs is averaged separately, with the \textbf{best} and \underline{second-best} values separately.}
    \vspace{-0.7em}
    \begin{adjustbox}{width=0.70\width,center}
    \begin{tabular}{c|c|cccccccccc}
    \toprule
    \rowcolor{mygray}
                             &                     & \multicolumn{5}{c|}{\textbf{Simulation}~(\textit{Attacked on OpenVLA-LIBERO-10})}                                                                                                           & \multicolumn{5}{c}{\textbf{Physical}~(\textit{Attacked on OpenVLA-7b})}                                                                                                             \\
\rowcolor{mygray}
\multirow{-2}{*}{\textbf{Targeted DoF(s)}}          & \multirow{-2}{*}{\textbf{Metric}} & \textbf{Spatial$^\triangle$} & \textbf{Object$^\triangle$} & \textbf{Goal$^\triangle$} & \textbf{Long$^*$} & \multicolumn{1}{c|}{\textbf{Avg}} & \textbf{Spatial$^\triangle$} & \textbf{Object$^\triangle$} & \textbf{Goal$^\triangle$} & \textbf{Long$^\triangle$} & \textbf{Avg} \\
\midrule
\midrule
\multicolumn{1}{c|}{Benign}               & \multirow{2}{*}{FR}                 & 15.3$\pm$10.2\%                           & 11.6$\pm$10.0\%                           & 20.8$\pm$12.0\%                         & 46.3$\pm$18.6\%                         & \multicolumn{1}{c|}{\cellcolor{mygray}23.5\%}    & 15.3$\pm$10.2\%                            & 11.6$\pm$10.0\%                           & 20.8$\pm$12.0\%                         & 46.3$\pm$18.6\%                         & \cellcolor{mygray} 23.5\%                        \\
\multicolumn{1}{c|}{Random Noise}         &                                     & 28.8$\pm$24.2\%                           & 14.8$\pm$7.9\%                            & 21.0$\pm$15.5\%                         & 48.4$\pm$14.8\%                         & \multicolumn{1}{c|}{\cellcolor{mygray}28.3\%}    & 28.8$\pm$24.2\%                            & 14.8$\pm$7.9\%                            & 21.0$\pm$15.5\%                         & 48.4$\pm$14.8\%                         & \cellcolor{mygray} 28.3\%                        \\
\midrule
\multicolumn{1}{c|}{DoF$_1$}              & \multirow{9}{*}{FR}                 & 100$\pm$0.0\%                           & 97.8$\pm$3.0\%                           & 100$\pm$0.0\%                         & 100$\pm$0.0\%                         & \multicolumn{1}{c|}{\cellcolor{mygray}99.5\%}                & 90.6$\pm$19.9\%                            & 36.6$\pm$18.1\%                            & 61.0$\pm$21.7\%                         & 86.4$\pm$12.7\%                         & \cellcolor{mygray} 68.7\%                        \\
\multicolumn{1}{c|}{DoF$_2$}              &                                     & 100$\pm$0.0\%                           & 99.2$\pm$1.3\%                           & 100$\pm$0.0\%                         & 100$\pm$0.0\%                         & \multicolumn{1}{c|}{\cellcolor{mygray}\underline{99.8\%}}    & 99.6$\pm$0.8\%                             & 69.2$\pm$11.1\%                            & 76.2$\pm$26.1\%                         & 91.8$\pm$9.9\%                          & \cellcolor{mygray} 84.2\%                        \\
\multicolumn{1}{c|}{DoF$_3$}              &                                     & 100$\pm$0.0\%                           & 97.2$\pm$6.0\%                           & 100$\pm$0.0\%                         & 100$\pm$0.0\%                         & \multicolumn{1}{c|}{\cellcolor{mygray}99.3\%}                & 72.0$\pm$14.8\%                            & 30.2$\pm$17.4\%                            & 39.2$\pm$20.9\%                         & 62.4$\pm$17.5\%                         & \cellcolor{mygray} 51.0\%                        \\
\multicolumn{1}{c|}{DoF$_4$}              &                                     & 94.0$\pm$11.0\%                         & 37.4$\pm$23.2\%                          & 51.2$\pm$26.5\%                       & 87.6$\pm$14.9\%                       & \multicolumn{1}{c|}{\cellcolor{mygray}67.6\%}                & 93.6$\pm$9.7\%                             & 31.8$\pm$20.3\%                            & 45.0$\pm$18.3\%                         & 63.8$\pm$17.0\%                         & \cellcolor{mygray} 58.6\%                        \\
\multicolumn{1}{c|}{DoF$_5$}              &                                     & 100$\pm$0.0\%                           & 92.6$\pm$8.4\%                           & 100$\pm$0.0\%                         & 100$\pm$0.0\%                         & \multicolumn{1}{c|}{\cellcolor{mygray}98.2\%}                & 97.8$\pm$4.9\%                             & 36.0$\pm$22.5\%                            & 62.0$\pm$24.8\%                         & 79.4$\pm$21.3\%                         & \cellcolor{mygray} 68.8\%                        \\
\multicolumn{1}{c|}{DoF$_6$}              &                                     & 100$\pm$0.0\%                           & 80.0$\pm$9.5\%                           & 92.6$\pm$13.4\%                       & 100$\pm$0.0\%                         & \multicolumn{1}{c|}{\cellcolor{mygray}93.2\%}                & 79.0$\pm$14.2\%                            & 22.2$\pm$14.4\%                            & 41.0$\pm$21.5\%                         & 62.6$\pm$15.7\%                         & \cellcolor{mygray} 51.2\%                        \\
\multicolumn{1}{c|}{DoF$_7$}              &                                     & 99.2$\pm$1.3\%                          & 86.6$\pm$8.3\%                           & 67.2$\pm$22.5\%                       & 96.4$\pm$6.4\%                        & \multicolumn{1}{c|}{\cellcolor{mygray}87.4\%}                & 91.8$\pm$15.3\%                            & 98.8$\pm$1.3\%                             & 70.6$\pm$23.4\%                         & 93.2$\pm$10.2\%                         & \cellcolor{mygray} \textbf{88.6\%}                        \\
\multicolumn{1}{c|}{\quad DoF$_{1\sim3}$} &                                     & 100$\pm$0.0\%                           & 100$\pm$0.0\%                           & 100$\pm$0.0\%                         & 100$\pm$0.0\%                         & \multicolumn{1}{c|}{\cellcolor{mygray}\textbf{100\%}}        & 96.4$\pm$8.0\%                             & 83.6$\pm$16.8\%                            & 74.4$\pm$26.3\%                         & 91.8$\pm$11.9\%                         & \cellcolor{mygray} \underline{86.6\%}                        \\
\midrule
\multicolumn{2}{c|}{Task Avg}                                                    & \cellcolor{mygray} \textbf{99.2\%}                 & \cellcolor{mygray} 86.4\%                 & \cellcolor{mygray} 88.9\%               & \cellcolor{mygray} \underline{98.0\%}               & \multicolumn{1}{c|}{-}         & \cellcolor{mygray} \textbf{90.1\%}                  & \cellcolor{mygray} 51.1\%                 & \cellcolor{mygray} 58.7\%               & \cellcolor{mygray} \underline{78.9\%}               & -                                                \\
\bottomrule
    \end{tabular}
    \label{tab:main-result-NTA}
    \end{adjustbox}
    \vspace{-15pt}
\end{table*}

\vspace{-2pt}
\subsection{Main Result}\label{subsec:main_results}
\vspace{-2pt}
\noindent \textbf{Quantitative Results.} 
For UADA and UPA, our methods effectively amplify action discrepancies, leading to a notable transfer attack ability in increasing failure rates~(see Tab.~\ref{table:simulation-results}).
Specifically, while attacking DoF$_1$ and DoF$_{1\sim3}$ in the \textbf{Simulation} setup, UADA and UPA achieve NAD of 21.0\% and 14.5\%, significantly outperforming UMA scenarios with increments of 6.9\% and 3.1\%, respectively. Both UADA and UPA effectively disrupt robot execution, yielding maximum average failure rates of 100\% and 89.7\%, respectively. 
Regarding the \textbf{Physical} setup~(see Tab.~\ref{table:physical-results}), our attack methods demonstrate strong transferability, as adversarial patches generated in the physical setting significantly impact the performance of LIBERO-X test. Specifically, both UADA and UPA generate malicious actions with large action discrepancies, which are 32.6\%~(DoF$_1$) and 26.9\%~(DoF$_{1\sim3}$) with an increase of 5.9\% and 3.1\% compared to 26.7\%~(DoF$_1$) and 23.8\%~(DoF$_{1\sim3}$) in UMA scenarios, respectively. \textbf{Compare the two settings}~(Tab.~\ref{table:simulation-results} $v.s.$ Tab.~\ref{table:physical-results}), we observe a general variance in NAD~(\eg32.6\% $v.s.$ 21.0\% for UADA). This discrepancy can be attributed to the fundamental differences between the simulation and real-world datasets used for conducting attacks~(Bridge V2~\cite{bridgev2} $v.s.$ LIBERO~\cite{LIBREO}). The increased variability in real-world data, including environmental complexity, object diversity, and task difficulty, allows the robot more opportunities to generate larger action discrepancies within the validation dataset.

For TMA task, we evaluate the effectiveness of our method by manipulating all DoF(s) to 0~(\ie, $y^{i}_T=0$ in Eq.~\ref{eq_tma}). Our method demonstrates significant effectiveness in manipulating robotic trajectory and increasing FR~(see Tab.~\ref{tab:main-result-NTA}). Specifically, when applied to the different generation settings, our approach yields notable increments across various tasks, including a max average failure rate 100\% $v.s.$ 23.5\% of benign performance in \textbf{Simulation}, and 88.6\% in \textbf{Physical}, respectively.
Moreover, during attack DoF$_4$, we observe low FR and high task deviation. This failure can be attributed to the fact that DoF$_4$ controls the orientation along the x-axis, which can be redundant DoF in tasks. As a result, attacks targeting \textit{redundant DoF(s)}~\cite{kurdila2019dynamics} are less likely to disrupt execution effectively. This observation underscores the importance of task-specific considerations when designing adversarial attacks on robotic systems.

\noindent \textbf{Qualitative Results.} 
We qualitatively analyze robot movement trajectories under the three proposed attacks in  Fig.~\ref{fig:qualitative}. 
For \textbf{UADA}, we compare the trajectory deviations in the same trail with patches generated from UADA and UMA attacks. As seen, the UMA induces small deviations in the trajectory. UADA, on the other hand, produces significantly larger trajectory deviations~(also supported by NAD metric in Tab.~\ref{table:physical-results}), which is attributed to UADA’s capability of incorporating action discrepancies. In our observation, UADA significantly amplifies the effect on overall task execution, thereby increasing the potential for robotic hazards.
For \textbf{UPA}, we observe chaotic and irregular behaviors, including instances where the end-effector moves out of the camera’s field of view. We attribute this phenomenon to the efficacy of the adversarial patch in perturbing the model’s spatial perception, inducing a consistent deviation from the intended direction of movement, ultimately resulting in failure in a long run.
For \textbf{TMA}, we observe a notable reduction in the range of motion along the x-axis corresponding to the targeted attack axis. This suggests that our proposed targeted attack can effectively constrained robot movements.

In summary, the qualitative analysis shows that both Untarget attack, UADA, UPA, and TMA can effectively disrupt the robot actions generated from VLA models. 
These findings underscore a \textit{pressing security concern} during the deployment of generalist robots, especially when considering application scenes that require reliable operations \cite{cheng2011reliability,Vogel2013safezone}.

\begin{figure}[htbp]
\vspace{-0.6em}
    \centering
    \includegraphics[width=0.88\linewidth]{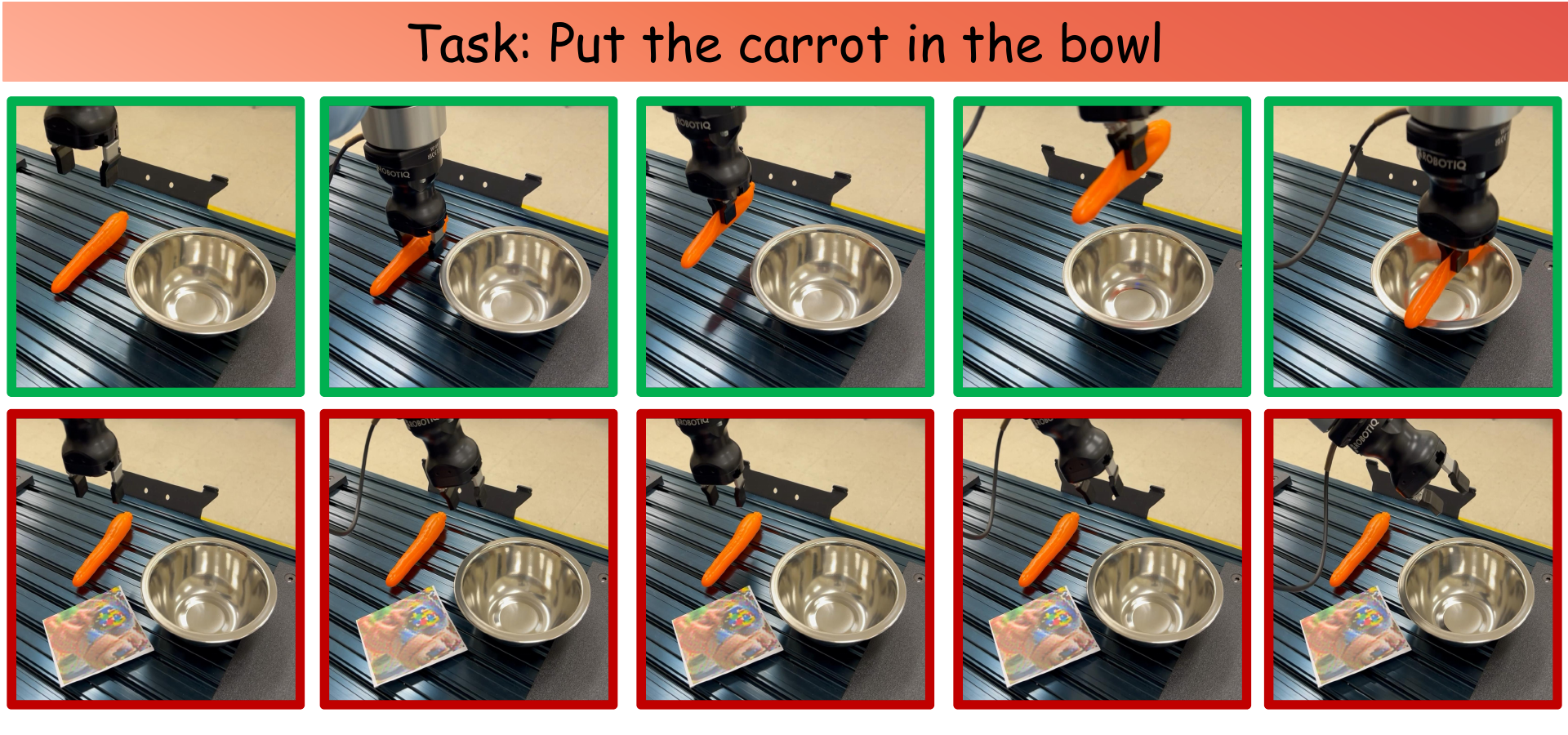}
    \vspace{-10pt}
    \caption{\textbf{Qualitative Results} of the physical world. The first/second row show \textcolor[HTML]{00B050}{benign} and \textcolor[HTML]{C00000}{adversarial} cases respectively.}
    \label{fig:physcial}
    \vspace{-1.2em}
\end{figure}

\begin{figure*}[htbp]
    \centering
    \includegraphics[width=0.95\linewidth]{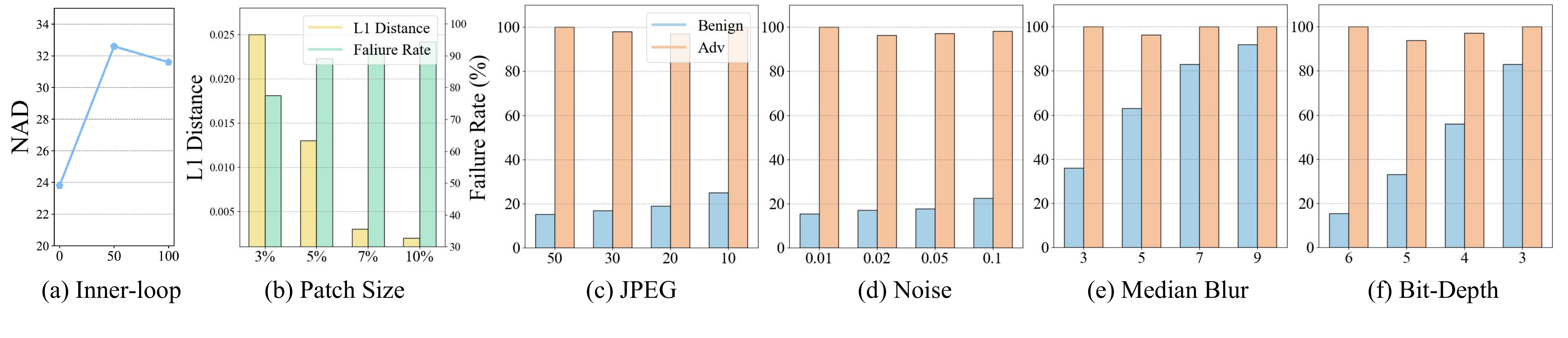}
    \vspace{-18pt}
    \caption{\textbf{Impact of Inner-loop, Patch Size and Defense Discussion.} The figure shows how varying
    Inner-loop affects NAD in UADA, and patch sizes affect $L1$ distance and the failure rates in TMA, both targeting at DoF$_1$. (a) Impact of Inner-loop, (b) Impact of Patch Size and (c-f) the effect of four different defenses on failure rates.}
    \label{fig:patchsize}
    \vspace{-18pt}
\end{figure*}

\noindent \textbf{Real-world Performance.} 
In addition to the digital simulation results, we also conduct a comprehensive evaluation of the performance of our generated adversarial patches in real-world scenarios. The evaluation encompassed 100 trials across three distinct tasks: object grasping, placement, and manipulation. As shown in Fig.~\ref{fig:physcial}, the adversarial patch generated with UADA demonstrated the ability to manipulate the robot effectively, achieving an attack success rate exceeding \textbf{43\%}. Although this success rate is lower than the corresponding digital-world performance (\ie, \textbf{100\%}), it highlights the effectiveness of our patches in physical-world applications as well without the need for further adaptations.
Crucially, we observed that the induced erratic movements~(similar to simulation scenario) of the robot during successful attacks pose significant risks to human safety and surrounding environments. This finding emphasizes the severe threat posed by VLA models in real-world applications, consistent with similar observations in digital settings.

\subsection{Diagnostic Experiment}\label{sec:diag}

\noindent \textbf{Impact of Inner-loops.} 
We discuss the impact of inner-loop steps to model performance in Fig.~\ref{fig:patchsize}(a). The results show that NAD first improves when inner-loop steps continue to increase. This is due to the reduced data variance and gradually stabilized gradients. However, further increasing the steps results in a slower performance increase, with a noticeably longer training schedule. We thus choose inner-loop steps=50 as it balances between performance and scale. \\
\noindent \textbf{Impact of Patch Size.} 
We study TMA’s performance with different patch sizes and report the average FR across four LIBERO~\cite{LIBREO} tasks in Fig.~\ref{fig:patchsize}(b). The patch sizes examined are $[3\%, 5\%, 7\%, 10\%]$ of input image.
Our findings indicate an inversely proportional relationship between the $L_1$ distance of predicted actions and the target action as the patch size increased. This increase in patch size also correlated with a marked rise in FR. The observed trend suggests larger patches give adversaries more optimization space to influence the model, aligning with prior work~\cite{cheng2024self,Zhang2022patchsize}.

\subsection{Robustness Evaluation} \label{sec:defense}

We further examine whether concurrent defense strategies~\cite{Gintare2016jpeg,Zhang2019noise,Xu2018medianblur} can resist our adversarial examples within VLA. Specifically, we applied four prior defense techniques and reported their FR for both benign and adversarial samples in Fig.~\ref{fig:patchsize}(c-f), respectively. The results show that our adversarial attack bypasses most defense strategies.

\subsection{Systemic Discussion} \label{sec:sysdis}
\begin{figure}[t]
    \centering
    \begin{subfigure}{0.23\textwidth}
        \includegraphics[width=\linewidth]{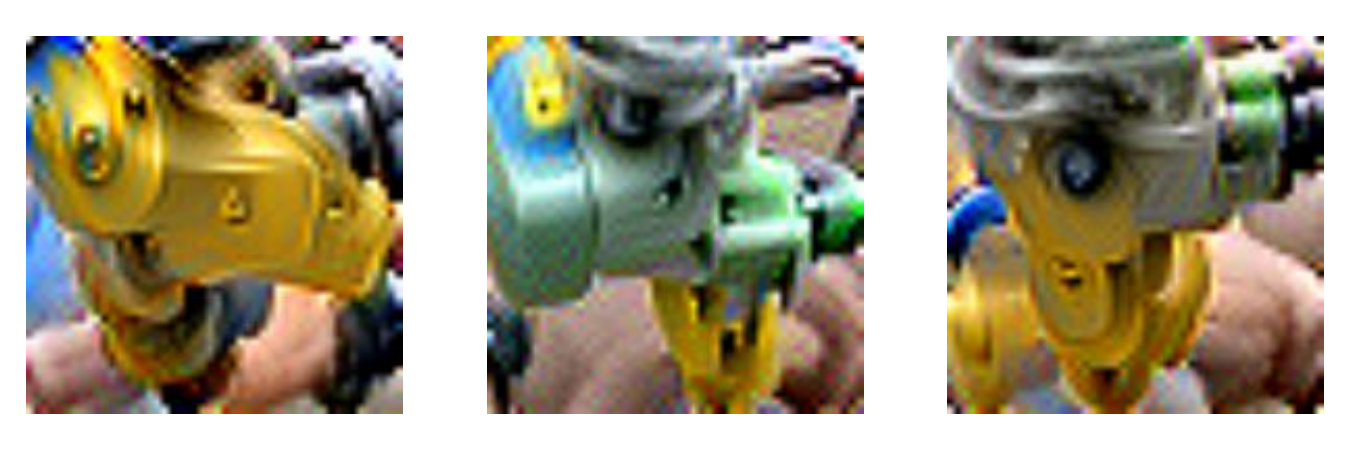}
        \caption{BridgeData V2~\cite{bridgev2}}
        \label{fig:vis1}
    \end{subfigure}
    \begin{subfigure}{0.23\textwidth}
        \includegraphics[width=\linewidth]{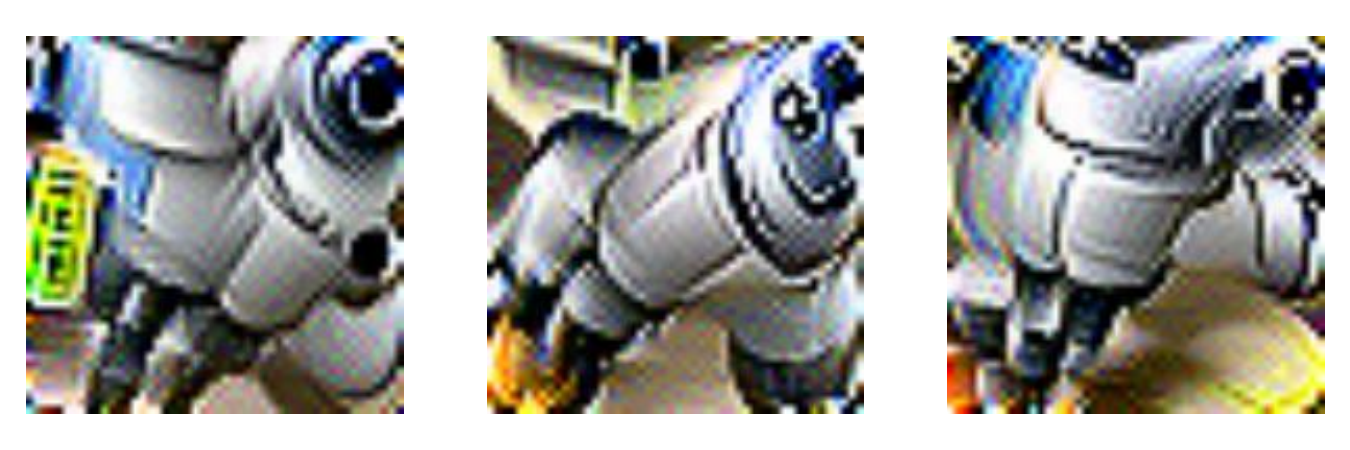}
        \caption{LIBERO~\cite{LIBREO}}
        \label{fig:vis2}
    \end{subfigure}
    \vspace{-8pt}
    \caption{\textbf{Patch Visualization.} Sematic-rich patches.}
    \label{fig:defensediscussion}
    \vspace{-1.5em}
\end{figure}
\noindent \textbf{Patch Pattern Analysis.} 
In Fig.~\ref{fig:defensediscussion}, we present the adversarial patches generated for a range of targets. A particularly noteworthy observation is that some of these patches bear a striking resemblance to the structural joints of a robotic arm. Given the strong similarity between these patches and robotic arms in appearance, we hypothesize that these VLA-based models, in their current training paradigms, are often limited to a narrow range of robotic systems operating within restricted camera views. This constrained perspective camera pose may inadvertently induce a learned representation bias, where adversarial perturbations align with prominent structural features of the training robot's appearance in order to deceive current victim models. 
Given that these patches are model-specific and reflect the spatial and visual patterns of the training environment, we have a reasonable concern that they may compromise VLA-based models' generalizability and robustness in physical-world deployments. \\
\noindent \textbf{Possible Defense Strategies.}
Our work highlights the pressing need for a paradigm shift in the VLA-based models' training strategies in order to successfully defend our attacks. Specifically, future approaches might incorporate more training scenarios that involve complex multi-robot interactions to mitigate the current patch bias. Another possible solution is to leverage the logical relationships within the robot arm’s physical structure to estimate the actual position of each joint. By predicting joint positions deviated from physical constraints, we can eliminate unreasonable adversarial samples, even if they are similar to actual joint(s).

\section{Conclusion}
While VLA models have gained significant popularity due to their substantial robot capabilities, this study pioneers specific robotic attack objectives and demonstrates that these models are in fact vulnerable to various types of attacks. Experimental results demonstrate that our attacks expose significant vulnerabilities to VLA models, raising concerns regarding impetuous real-world deployments. We believe our framework provides pioneering and foundational contributions to the reliability of AI-enhanced robotic systems.

\section*{Acknowledgements}
This research was supported by the National Science Foundation under Grant No.~2450068 and No.~2242243, as well as start-up resources provided by Rutgers University. The views and conclusions contained herein are those of the authors and should not be interpreted as necessarily representing the official policies or endorsements, either expressed or implied, of U.S. Naval Research Laboratory (NRL) or the U.S. Government. The U.S. Government is authorized to reproduce and distribute reprints for Government purposes notwithstanding any copyright notation herein.
{
    \small
    \bibliographystyle{ieeenat_fullname}
    \bibliography{ref}

\begin{thebibliography}{72}
\providecommand{\natexlab}[1]{#1}
\providecommand{\url}[1]{\texttt{#1}}
\expandafter\ifx\csname urlstyle\endcsname\relax
  \providecommand{\doi}[1]{doi: #1}\else
  \providecommand{\doi}{doi: \begingroup \urlstyle{rm}\Url}\fi

\bibitem[Al-Hamadani et~al.(2024)Al-Hamadani, Fadhel, Alzubaidi, and Harangi]{al2024reinforcement}
Mokhaled~NA Al-Hamadani, Mohammed~A Fadhel, Laith Alzubaidi, and Balazs Harangi.
\newblock Reinforcement learning algorithms and applications in healthcare and robotics: A comprehensive and systematic review.
\newblock \emph{Sensors}, 24\penalty0 (8):\penalty0 2461, 2024.

\bibitem[Athalye et~al.(2018{\natexlab{a}})Athalye, Carlini, and Wagner]{AthalyeC018Obfuscated}
Anish Athalye, Nicholas Carlini, and David~A. Wagner.
\newblock Obfuscated gradients give a false sense of security: Circumventing defenses to adversarial examples.
\newblock In \emph{ICML}, 2018{\natexlab{a}}.

\bibitem[Athalye et~al.(2018{\natexlab{b}})Athalye, Engstrom, Ilyas, and Kwok]{AthalyeEOT2018}
Anish Athalye, Logan Engstrom, Andrew Ilyas, and Kevin Kwok.
\newblock Synthesizing robust adversarial examples.
\newblock In \emph{ICML}, 2018{\natexlab{b}}.

\bibitem[Billard and Kragic(2019)]{BillardscienceTrends}
Aude Billard and Danica Kragic.
\newblock Trends and challenges in robot manipulation.
\newblock \emph{Science}, 364\penalty0 (6446), 2019.

\bibitem[Brohan et~al.(2023)Brohan, Brown, Carbajal, Chebotar, Dabis, Finn, Gopalakrishnan, Hausman, Herzog, Hsu, Ibarz, Ichter, Irpan, Jackson, Jesmonth, Joshi, Julian, Kalashnikov, Kuang, Leal, Lee, Levine, Lu, Malla, Manjunath, Mordatch, Nachum, Parada, Peralta, Perez, Pertsch, Quiambao, Rao, Ryoo, Salazar, Sanketi, Sayed, Singh, Sontakke, Stone, Tan, Tran, Vanhoucke, Vega, Vuong, Xia, Xiao, Xu, Xu, Yu, and Zitkovich]{Brohan2023rt1}
Anthony Brohan, Noah Brown, Justice Carbajal, Yevgen Chebotar, Joseph Dabis, Chelsea Finn, Keerthana Gopalakrishnan, Karol Hausman, Alexander Herzog, Jasmine Hsu, Julian Ibarz, Brian Ichter, Alex Irpan, Tomas Jackson, Sally Jesmonth, Nikhil~J. Joshi, Ryan Julian, Dmitry Kalashnikov, Yuheng Kuang, Isabel Leal, Kuang{-}Huei Lee, Sergey Levine, Yao Lu, Utsav Malla, Deeksha Manjunath, Igor Mordatch, Ofir Nachum, Carolina Parada, Jodilyn Peralta, Emily Perez, Karl Pertsch, Jornell Quiambao, Kanishka Rao, Michael~S. Ryoo, Grecia Salazar, Pannag~R. Sanketi, Kevin Sayed, Jaspiar Singh, Sumedh Sontakke, Austin Stone, Clayton Tan, Huong~T. Tran, Vincent Vanhoucke, Steve Vega, Quan Vuong, Fei Xia, Ted Xiao, Peng Xu, Sichun Xu, Tianhe Yu, and Brianna Zitkovich.
\newblock {RT-1:} robotics transformer for real-world control at scale.
\newblock In \emph{Robotics: Science and Systems XIX}, 2023.

\bibitem[Brown et~al.(2018)Brown, Mané, Roy, Abadi, and mer]{brown2018adversarialpatch}
Tom~B. Brown, Dandelion Mané, Aurko Roy, Martín Abadi, and Justin~Gil mer.
\newblock Adversarial patch.
\newblock \emph{arXiv preprint arXiv:1712.09665}, 2018.

\bibitem[Carlini and Wagner(2017)]{Carlini017Bypassing}
Nicholas Carlini and David~A. Wagner.
\newblock Adversarial examples are not easily detected: Bypassing ten detection methods.
\newblock In \emph{Proceedings of the 10th {ACM} Workshop on Artificial Intelligence and Security}, pages 3--14, 2017.

\bibitem[Chen et~al.(2023)Chen, Wang, Beyer, Kolesnikov, Wu, Voigtlaender, Mustafa, Goodman, Alabdulmohsin, Padlewski, et~al.]{chen2023pali}
Xi Chen, Xiao Wang, Lucas Beyer, Alexander Kolesnikov, Jialin Wu, Paul Voigtlaender, Basil Mustafa, Sebastian Goodman, Ibrahim Alabdulmohsin, Piotr Padlewski, et~al.
\newblock Pali-3 vision language models: Smaller, faster, stronger.
\newblock \emph{arXiv preprint arXiv:2310.09199}, 2023.

\bibitem[Cheng and Dhillon(2011)]{cheng2011reliability}
Shen Cheng and BS Dhillon.
\newblock Reliability and availability analysis of a robot-safety system.
\newblock \emph{Journal of Quality in Maintenance Engineering}, 17:\penalty0 203--232, 2011.

\bibitem[Cheng et~al.(2024)Cheng, Han, Liang, Wang, Zhang, and Liu]{cheng2024self}
Zhiyuan Cheng, Cheng Han, James Liang, Qifan Wang, Xiangyu Zhang, and Dongfang Liu.
\newblock Self-supervised adversarial training of monocular depth estimation against physical-world attacks.
\newblock \emph{arXiv preprint arXiv:2406.05857}, 2024.

\bibitem[Chiang et~al.(2024)Chiang, Xu, Fu, Jacob, Zhang, Lee, Yu, Schenck, Rendleman, Shah, et~al.]{chiang2024mobility}
Hao-Tien~Lewis Chiang, Zhuo Xu, Zipeng Fu, Mithun~George Jacob, Tingnan Zhang, Tsang-Wei~Edward Lee, Wenhao Yu, Connor Schenck, David Rendleman, Dhruv Shah, et~al.
\newblock Mobility vla: Multimodal instruction navigation with long-context vlms and topological graphs.
\newblock \emph{arXiv preprint arXiv:2407.07775}, 2024.

\bibitem[Cui and Trinkle(2021)]{Cui2021ScienceRobotic}
Jinda Cui and Jeff Trinkle.
\newblock Toward next-generation learned robot manipulation.
\newblock \emph{Sci. Robotics}, 6\penalty0 (54):\penalty0 9461, 2021.

\bibitem[Cui et~al.(2024)Cui, Han, and Liu]{cui2024collaborative}
Yiming Cui, Cheng Han, and Dongfang Liu.
\newblock Collaborative multi-task learning for multi-object tracking and segmentation.
\newblock \emph{Journal on Autonomous Transportation Systems}, 1:\penalty0 1--23, 2024.

\bibitem[Ding et~al.(2025)Ding, Zhao, Zhang, Song, Zhang, Huang, Yang, and Wang]{ding2025quar}
Pengxiang Ding, Han Zhao, Wenjie Zhang, Wenxuan Song, Min Zhang, Siteng Huang, Ningxi Yang, and Donglin Wang.
\newblock Quar-vla: Vision-language-action model for quadruped robots.
\newblock In \emph{ECCV}, 2025.

\bibitem[Du et~al.(2022)Du, Chen, Chin, Law, Sasdelli, Rajasegaran, and Campbell]{Du2022Physical}
Andrew Du, Bo Chen, Tat{-}Jun Chin, Yee~Wei Law, Michele Sasdelli, Ramesh Rajasegaran, and Dillon Campbell.
\newblock Physical adversarial attacks on an aerial imagery object detector.
\newblock In \emph{WACV}, 2022.

\bibitem[Dziugaite et~al.(2016)Dziugaite, Ghahramani, and Roy]{Gintare2016jpeg}
Gintare~Karolina Dziugaite, Zoubin Ghahramani, and Daniel~M. Roy.
\newblock A study of the effect of {JPG} compression on adversarial images.
\newblock \emph{arXiv preprint arXiv:1608.00853}, 2016.

\bibitem[Finn et~al.(2017)Finn, Yu, Zhang, Abbeel, and Levine]{Finn2017MetaLearning}
Chelsea Finn, Tianhe Yu, Tianhao Zhang, Pieter Abbeel, and Sergey Levine.
\newblock One-shot visual imitation learning via meta-learning.
\newblock In \emph{CoRL}, 2017.

\bibitem[Goodfellow et~al.(2015)Goodfellow, Shlens, and Szegedy]{Goodfellow2015FGSM}
Ian~J. Goodfellow, Jonathon Shlens, and Christian Szegedy.
\newblock Explaining and harnessing adversarial examples.
\newblock In \emph{ICLR}, 2015.

\bibitem[Haddadin et~al.(2022)Haddadin, Parusel, Johannsmeier, Golz, Gabl, Walch, Sabaghian, J{\"a}hne, Hausperger, and Haddadin]{haddadin2022franka}
Sami Haddadin, Sven Parusel, Lars Johannsmeier, Saskia Golz, Simon Gabl, Florian Walch, Mohamadreza Sabaghian, Christoph J{\"a}hne, Lukas Hausperger, and Simon Haddadin.
\newblock The franka emika robot: A reference platform for robotics research and education.
\newblock \emph{IEEE Robotics \& Automation Magazine}, 29:\penalty0 46--64, 2022.

\bibitem[Huang et~al.(2023{\natexlab{a}})Huang, Yong, Ma, Linghu, Li, Wang, Li, Zhu, Jia, and Huang]{huang2023embodied}
Jiangyong Huang, Silong Yong, Xiaojian Ma, Xiongkun Linghu, Puhao Li, Yan Wang, Qing Li, Song-Chun Zhu, Baoxiong Jia, and Siyuan Huang.
\newblock An embodied generalist agent in 3d world.
\newblock \emph{arXiv preprint arXiv:2311.12871}, 2023{\natexlab{a}}.

\bibitem[Huang et~al.(2023{\natexlab{b}})Huang, Wang, Zhang, Li, Wu, and Fei{-}Fei]{Huang2023VoxPoser}
Wenlong Huang, Chen Wang, Ruohan Zhang, Yunzhu Li, Jiajun Wu, and Li Fei{-}Fei.
\newblock Voxposer: Composable 3d value maps for robotic manipulation with language models.
\newblock In \emph{CoRL}, 2023{\natexlab{b}}.

\bibitem[Ingrand and Ghallab(2017)]{Ingrand2017ReviewDeliberation}
F{\'{e}}lix Ingrand and Malik Ghallab.
\newblock Deliberation for autonomous robots: {A} survey.
\newblock \emph{Artif. Intell.}, 247:\penalty0 10--44, 2017.

\bibitem[ISO 10218-1/2:2011()]{iso_10218}
ISO 10218-1/2:2011.
\newblock {Robots and Robotic Devices Safety Requirements for Industrial Robots Part 1: Robots/Part 2: Robot Systems and Integration}.
\newblock Standard, International Organization for Standardization, 2011.

\bibitem[ISO13482(2014)]{ISO13482}
ISO13482.
\newblock Robots and robotic devices — safety requirements for personal care robots.
\newblock \emph{International Organization for Standardization}, 2014.

\bibitem[ISO/TS 15066:2016()]{iso_15066}
ISO/TS 15066:2016.
\newblock {Robots and Robotic Devices Collaborative Robots}.
\newblock Standard, International Organization for Standardization, 2016.

\bibitem[James et~al.(2018)James, Bloesch, and Davison]{JamesBD18TaskEmbedded}
Stephen James, Michael Bloesch, and Andrew~J. Davison.
\newblock Task-embedded control networks for few-shot imitation learning.
\newblock In \emph{CoRL}, 2018.

\bibitem[Jiang et~al.(2022)Jiang, Gupta, Zhang, Wang, Dou, Chen, Fei{-}Fei, Anandkumar, Zhu, and Fan]{Jiang2022VIMA}
Yunfan Jiang, Agrim Gupta, Zichen Zhang, Guanzhi Wang, Yongqiang Dou, Yanjun Chen, Li Fei{-}Fei, Anima Anandkumar, Yuke Zhu, and Linxi Fan.
\newblock {VIMA:} general robot manipulation with multimodal prompts.
\newblock \emph{arXiv preprint arXiv:2210.03094}, 2022.

\bibitem[Karamcheti et~al.(2024)Karamcheti, Nair, Balakrishna, Liang, Kollar, and Sadigh]{karamcheti2024prismatic}
Siddharth Karamcheti, Suraj Nair, Ashwin Balakrishna, Percy Liang, Thomas Kollar, and Dorsa Sadigh.
\newblock Prismatic vlms: Investigating the design space of visually-conditioned language models.
\newblock \emph{arXiv preprint arXiv:2402.07865}, 2024.

\bibitem[Kim et~al.(2024)Kim, Pertsch, Karamcheti, Xiao, Balakrishna, Nair, Rafailov, Foster, Lam, Sanketi, Vuong, Kollar, Burchfiel, Tedrake, Sadigh, Levine, Liang, and Finn]{kim24openvla}
{Moo Jin} Kim, Karl Pertsch, Siddharth Karamcheti, Ted Xiao, Ashwin Balakrishna, Suraj Nair, Rafael Rafailov, Ethan Foster, Grace Lam, Pannag Sanketi, Quan Vuong, Thomas Kollar, Benjamin Burchfiel, Russ Tedrake, Dorsa Sadigh, Sergey Levine, Percy Liang, and Chelsea Finn.
\newblock Openvla: An open-source vision-language-action model.
\newblock \emph{arXiv preprint arXiv:2406.09246}, 2024.

\bibitem[Kroemer et~al.(2021)Kroemer, Niekum, and Konidaris]{Kroemer2021ReviewRobotLearning}
Oliver Kroemer, Scott Niekum, and George Konidaris.
\newblock A review of robot learning for manipulation: Challenges, representations, and algorithms.
\newblock \emph{J. Mach. Learn. Res.}, 22:\penalty0 30:1--30:82, 2021.

\bibitem[Kurdila and Ben-Tzvi(2019)]{kurdila2019dynamics}
Andrew~J Kurdila and Pinhas Ben-Tzvi.
\newblock \emph{Dynamics and control of robotic systems}.
\newblock John Wiley \& Sons, 2019.

\bibitem[Levine et~al.(2018)Levine, Pastor, Krizhevsky, Ibarz, and Quillen]{Levine2018CNNHandEye}
Sergey Levine, Peter Pastor, Alex Krizhevsky, Julian Ibarz, and Deirdre Quillen.
\newblock Learning hand-eye coordination for robotic grasping with deep learning and large-scale data collection.
\newblock \emph{Int. J. Robotics Res.}, 37\penalty0 (4-5):\penalty0 421--436, 2018.

\bibitem[Li et~al.(2023)Li, Liu, Zhang, Yu, Xu, Wu, Cheang, Jing, Zhang, Liu, et~al.]{li2023vision}
Xinghang Li, Minghuan Liu, Hanbo Zhang, Cunjun Yu, Jie Xu, Hongtao Wu, Chilam Cheang, Ya Jing, Weinan Zhang, Huaping Liu, et~al.
\newblock Vision-language foundation models as effective robot imitators.
\newblock \emph{arXiv preprint arXiv:2311.01378}, 2023.

\bibitem[Li et~al.(2024)Li, Mata, Park, Kahatapitiya, Jang, Shang, Ranasinghe, Burgert, Cai, Lee, and Ryoo]{Li2024LLaRA}
Xiang Li, Cristina Mata, Jongwoo Park, Kumara Kahatapitiya, Yoo~Sung Jang, Jinghuan Shang, Kanchana Ranasinghe, Ryan Burgert, Mu Cai, Yong~Jae Lee, and Michael~S. Ryoo.
\newblock Llara: Supercharging robot learning data for vision-language policy.
\newblock \emph{arXiv preprint arXiv:2406.20095}, 2024.

\bibitem[Liu et~al.(2019)Liu, Liu, Fan, Ma, Zhang, Xie, and Tao]{Liu2019Perceptualpatch}
Aishan Liu, Xianglong Liu, Jiaxin Fan, Yuqing Ma, Anlan Zhang, Huiyuan Xie, and Dacheng Tao.
\newblock Perceptual-sensitive {GAN} for generating adversarial patches.
\newblock In \emph{AAAI}, 2019.

\bibitem[Liu et~al.(2023)Liu, Zhu, Gao, Feng, Liu, Zhu, and Stone]{LIBREO}
Bo Liu, Yifeng Zhu, Chongkai Gao, Yihao Feng, Qiang Liu, Yuke Zhu, and Peter Stone.
\newblock {LIBERO:} benchmarking knowledge transfer for lifelong robot learning.
\newblock In \emph{NeurIPS}, 2023.

\bibitem[Liu et~al.(2024{\natexlab{a}})Liu, Li, Li, and Lee]{liu2024improved}
Haotian Liu, Chunyuan Li, Yuheng Li, and Yong~Jae Lee.
\newblock Improved baselines with visual instruction tuning.
\newblock In \emph{CVPR}, 2024{\natexlab{a}}.

\bibitem[Liu et~al.(2024{\natexlab{b}})Liu, Li, Wu, and Lee]{liu2024visual}
Haotian Liu, Chunyuan Li, Qingyang Wu, and Yong~Jae Lee.
\newblock Visual instruction tuning.
\newblock In \emph{NeurIPS}, 2024{\natexlab{b}}.

\bibitem[Madry et~al.(2018)Madry, Makelov, Schmidt, Tsipras, and Vladu]{Madry2018twords}
Aleksander Madry, Aleksandar Makelov, Ludwig Schmidt, Dimitris Tsipras, and Adrian Vladu.
\newblock Towards deep learning models resistant to adversarial attacks.
\newblock In \emph{ICLR}, 2018.

\bibitem[Mason(2001)]{Mason2001Mechanics}
Matthew~T. Mason.
\newblock \emph{Mechanics of Robotic Manipulation}.
\newblock {MIT} Press, 2001.

\bibitem[Mason(2018)]{Mason2018Toward}
Matthew~T. Mason.
\newblock Toward robotic manipulation.
\newblock \emph{Annu. Rev. Control. Robotics Auton. Syst.}, 1:\penalty0 1--28, 2018.

\bibitem[Mees et~al.(2022)Mees, Hermann, and Burgard]{MeesHB2022What}
Oier Mees, Luk{\'{a}}s Hermann, and Wolfram Burgard.
\newblock What matters in language conditioned robotic imitation learning over unstructured data.
\newblock \emph{{IEEE} Robotics Autom. Lett.}, 7\penalty0 (4):\penalty0 11205--11212, 2022.

\bibitem[Moudgal et~al.(1994)Moudgal, Passino, and Yurkovich]{moudgal1994rule}
Vivek~G Moudgal, Kevin~M Passino, and Stephen Yurkovich.
\newblock Rule-based control for a flexible-link robot.
\newblock \emph{IEEE Trans. on Control Sys. Tech.}, 2\penalty0 (4):\penalty0 392--405, 1994.

\bibitem[Murray et~al.(1994)Murray, Li, and Sastry]{Murray1994mathematical}
Richard~M. Murray, Zexiang Li, and S.~Shankar Sastry.
\newblock \emph{A mathematical introduction to robotics manipulation}.
\newblock {CRC} Press, 1994.

\bibitem[Pinto and Gupta(2016)]{Pinto2016Supersizing}
Lerrel Pinto and Abhinav Gupta.
\newblock Supersizing self-supervision: Learning to grasp from 50k tries and 700 robot hours.
\newblock In \emph{ICRA}, 2016.

\bibitem[Pong et~al.(2020)Pong, Dalal, Lin, Nair, Bahl, and Levine]{Pong2020Skew}
Vitchyr Pong, Murtaza Dalal, Steven Lin, Ashvin Nair, Shikhar Bahl, and Sergey Levine.
\newblock Skew-fit: State-covering self-supervised reinforcement learning.
\newblock In \emph{ICML}, 2020.

\bibitem[Qi et~al.(2024)Qi, Huang, Panda, Henderson, Wang, and Mittal]{QiHP0WM24Visual}
Xiangyu Qi, Kaixuan Huang, Ashwinee Panda, Peter Henderson, Mengdi Wang, and Prateek Mittal.
\newblock Visual adversarial examples jailbreak aligned large language models.
\newblock In \emph{AAAI}, 2024.

\bibitem[Sharif et~al.(2016)Sharif, Bhagavatula, Bauer, and Reiter]{Sharif2016advglass}
Mahmood Sharif, Sruti Bhagavatula, Lujo Bauer, and Michael~K. Reiter.
\newblock Accessorize to a crime: Real and stealthy attacks on state-of-the-art face recognition.
\newblock In \emph{ACM SIGSAC}, 2016.

\bibitem[Smith and Coles(1973)]{Nilsson1973}
Michael~H. Smith and L.~Stephen Coles.
\newblock Design of a low cost, general purpose robot.
\newblock In \emph{IJCAI}, 1973.

\bibitem[Su et~al.(2019)Su, Vargas, and Sakurai]{Su2019OnePixel}
Jiawei Su, Danilo~Vasconcellos Vargas, and Kouichi Sakurai.
\newblock One pixel attack for fooling deep neural networks.
\newblock \emph{{IEEE} Trans. Evol. Comput.}, 23\penalty0 (5):\penalty0 828--841, 2019.

\bibitem[Szegedy et~al.(2014)Szegedy, Zaremba, Sutskever, Bruna, Erhan, Goodfellow, and Fergus]{Szegedy2014Intriguing}
Christian Szegedy, Wojciech Zaremba, Ilya Sutskever, Joan Bruna, Dumitru Erhan, Ian~J. Goodfellow, and Rob Fergus.
\newblock Intriguing properties of neural networks.
\newblock In \emph{ICLR}, 2014.

\bibitem[Tang et~al.(2024)Tang, Abbatematteo, Hu, Chandra, Mart{\'\i}n-Mart{\'\i}n, and Stone]{tang2024deep}
Chen Tang, Ben Abbatematteo, Jiaheng Hu, Rohan Chandra, Roberto Mart{\'\i}n-Mart{\'\i}n, and Peter Stone.
\newblock Deep reinforcement learning for robotics: A survey of real-world successes.
\newblock \emph{arXiv preprint arXiv:2408.03539}, 2024.

\bibitem[Team et~al.(2024)Team, Ghosh, Walke, Pertsch, Black, Mees, Dasari, Hejna, Kreiman, Xu, Luo, Tan, Chen, Sanketi, Vuong, Xiao, Sadigh, Finn, and Levine]{Octo2024Octo}
Octo~Model Team, Dibya Ghosh, Homer Walke, Karl Pertsch, Kevin Black, Oier Mees, Sudeep Dasari, Joey Hejna, Tobias Kreiman, Charles Xu, Jianlan Luo, You~Liang Tan, Lawrence~Yunliang Chen, Pannag Sanketi, Quan Vuong, Ted Xiao, Dorsa Sadigh, Chelsea Finn, and Sergey Levine.
\newblock Octo: An open-source generalist robot policy.
\newblock \emph{arXiv preprint arXiv:2405.12213}, 2024.

\bibitem[Tram{\`{e}}r(2022)]{Tramer22Detecting}
Florian Tram{\`{e}}r.
\newblock Detecting adversarial examples is (nearly) as hard as classifying them.
\newblock In \emph{ICML}, 2022.

\bibitem[Tram{\`{e}}r et~al.(2020)Tram{\`{e}}r, Carlini, Brendel, and Madry]{Florian2020untarget}
Florian Tram{\`{e}}r, Nicholas Carlini, Wieland Brendel, and Aleksander Madry.
\newblock On adaptive attacks to adversarial example defenses.
\newblock In \emph{NeurIPS}, 2020.

\bibitem[Vogel et~al.(2013)Vogel, Walter, and Elkmann]{Vogel2013safezone}
Christian Vogel, Christoph Walter, and Norbert Elkmann.
\newblock A projection-based sensor system for safe physical human-robot collaboration.
\newblock In \emph{IROS}, 2013.

\bibitem[Walke et~al.(2023)Walke, Black, Zhao, Vuong, Zheng, Hansen{-}Estruch, He, Myers, Kim, Du, Lee, Fang, Finn, and Levine]{bridgev2}
Homer~Rich Walke, Kevin Black, Tony~Z. Zhao, Quan Vuong, Chongyi Zheng, Philippe Hansen{-}Estruch, Andre~Wang He, Vivek Myers, Moo~Jin Kim, Max Du, Abraham Lee, Kuan Fang, Chelsea Finn, and Sergey Levine.
\newblock Bridgedata {V2:} {A} dataset for robot learning at scale.
\newblock In \emph{CoRL}, 2023.

\bibitem[Wang et~al.(2024{\natexlab{a}})Wang, Liu, Liang, Cui, Mao, Nie, Liu, Feng, Xu, Han, et~al.]{wang2024mmpt}
Taowen Wang, Yiyang Liu, James~Chenhao Liang, Yiming Cui, Yuning Mao, Shaoliang Nie, Jiahao Liu, Fuli Feng, Zenglin Xu, Cheng Han, et~al.
\newblock {M\({}^{2}\)}pt: Multimodal prompt tuning for zero-shot instruction learning.
\newblock In \emph{EMNLP}, 2024{\natexlab{a}}.

\bibitem[Wang et~al.(2019)Wang, Zheng, Song, Wang, Rahimpour, and Qi]{Wang2019advPattern}
Zhibo Wang, Siyan Zheng, Mengkai Song, Qian Wang, Alireza Rahimpour, and Hairong Qi.
\newblock advpattern: Physical-world attacks on deep person re-identification via adversarially transformable patterns.
\newblock In \emph{ICCV}, 2019.

\bibitem[Wang et~al.(2024{\natexlab{b}})Wang, Yan, Wang, Xu, Wu, and Wang]{wang2024research}
Zixiang Wang, Hao Yan, Zhuoyue Wang, Zhengjia Xu, Zhizhong Wu, and Yining Wang.
\newblock Research on autonomous robots navigation based on reinforcement learning.
\newblock In \emph{RAIIC}, 2024{\natexlab{b}}.

\bibitem[Wang et~al.(2024{\natexlab{c}})Wang, Zhou, Song, Huang, Shu, and Ma]{wang2024towards}
Zhijie Wang, Zhehua Zhou, Jiayang Song, Yuheng Huang, Zhan Shu, and Lei Ma.
\newblock Towards testing and evaluating vision-language-action models for robotic manipulation: An empirical study.
\newblock \emph{arXiv preprint arXiv:2409.12894}, 2024{\natexlab{c}}.

\bibitem[Wen et~al.(2024)Wen, Zhu, Li, Zhu, Wu, Xu, Cheng, Shen, Peng, Feng, et~al.]{wen2024tinyvla}
Junjie Wen, Yichen Zhu, Jinming Li, Minjie Zhu, Kun Wu, Zhiyuan Xu, Ran Cheng, Chaomin Shen, Yaxin Peng, Feifei Feng, et~al.
\newblock Tinyvla: Towards fast, data-efficient vision-language-action models for robotic manipulation.
\newblock \emph{arXiv preprint arXiv:2409.12514}, 2024.

\bibitem[{Wikipedia contributors}(2024)]{enwiki-1249481595}
{Wikipedia contributors}.
\newblock Finch (film) --- {Wikipedia}{,} the free encyclopedia.
\newblock \url{https://en.wikipedia.org/w/index.php?title=Finch_(film)&oldid=1249481595}, 2024.
\newblock [Online; accessed 5-November-2024].

\bibitem[Wu et~al.(2024)Wu, Koh, Salakhutdinov, Fried, and Raghunathan]{wu2024adversarial}
Chen~Henry Wu, Jing~Yu Koh, Ruslan Salakhutdinov, Daniel Fried, and Aditi Raghunathan.
\newblock Adversarial attacks on multimodal agents.
\newblock \emph{arXiv preprint arXiv:2406.12814}, 2024.

\bibitem[Xu et~al.(2020)Xu, Zhang, Liu, Fan, Sun, Chen, Chen, Wang, and Lin]{Xu2020AdvTShirt}
Kaidi Xu, Gaoyuan Zhang, Sijia Liu, Quanfu Fan, Mengshu Sun, Hongge Chen, Pin{-}Yu Chen, Yanzhi Wang, and Xue Lin.
\newblock Adversarial t-shirt! evading person detectors in a physical world.
\newblock In \emph{ECCV}, 2020.

\bibitem[Xu et~al.(2018)Xu, Evans, and Qi]{Xu2018medianblur}
Weilin Xu, David Evans, and Yanjun Qi.
\newblock Feature squeezing: Detecting adversarial examples in deep neural networks.
\newblock In \emph{NDSS}, 2018.

\bibitem[Xu et~al.(2024)Xu, Feng, Shao, Ashby, Shen, Jin, Cheng, Wang, and Huang]{xu2024vision}
Zhiyang Xu, Chao Feng, Rulin Shao, Trevor Ashby, Ying Shen, Di Jin, Yu Cheng, Qifan Wang, and Lifu Huang.
\newblock Vision-flan: Scaling human-labeled tasks in visual instruction tuning.
\newblock \emph{arXiv preprint arXiv:2402.11690}, 2024.

\bibitem[Zhang and Liang(2019)]{Zhang2019noise}
Yuchen Zhang and Percy Liang.
\newblock Defending against whitebox adversarial attacks via randomized discretization.
\newblock In \emph{AISTATS}, 2019.

\bibitem[Zhang et~al.(2022)Zhang, Tan, Chen, Liu, Zhang, and Li]{Zhang2022patchsize}
Yaoyuan Zhang, Yu{-}an Tan, Tian Chen, Xinrui Liu, Quanxin Zhang, and Yuanzhang Li.
\newblock Enhancing the transferability of adversarial examples with random patch.
\newblock In \emph{IJCAI}, 2022.

\bibitem[Zhao et~al.(2023)Zhao, Wu, He, and Huang]{zhao2023svit}
Bo Zhao, Boya Wu, Muyang He, and Tiejun Huang.
\newblock Svit: Scaling up visual instruction tuning.
\newblock \emph{arXiv preprint arXiv:2307.04087}, 2023.

\bibitem[Zhu et~al.(2021)Zhu, Wang, Chen, and Dong]{zhu2021rule}
Yuanyang Zhu, Zhi Wang, Chunlin Chen, and Daoyi Dong.
\newblock Rule-based reinforcement learning for efficient robot navigation with space reduction.
\newblock \emph{IEEE/ASME Trans. on Mechatronics}, 27:\penalty0 846--857, 2021.

\bibitem[Zitkovich et~al.(2023)Zitkovich, Yu, Xu, Xu, Xiao, Xia, Wu, Wohlhart, Welker, Wahid, Vuong, Vanhoucke, Tran, Soricut, Singh, Singh, Sermanet, Sanketi, Salazar, Ryoo, Reymann, Rao, Pertsch, Mordatch, Michalewski, Lu, Levine, Lee, Lee, Leal, Kuang, Kalashnikov, Julian, Joshi, Irpan, Ichter, Hsu, Herzog, Hausman, Gopalakrishnan, Fu, Florence, Finn, Dubey, Driess, Ding, Choromanski, Chen, Chebotar, Carbajal, Brown, Brohan, Arenas, and Han]{Brianna2023RT2}
Brianna Zitkovich, Tianhe Yu, Sichun Xu, Peng Xu, Ted Xiao, Fei Xia, Jialin Wu, Paul Wohlhart, Stefan Welker, Ayzaan Wahid, Quan Vuong, Vincent Vanhoucke, Huong~T. Tran, Radu Soricut, Anikait Singh, Jaspiar Singh, Pierre Sermanet, Pannag~R. Sanketi, Grecia Salazar, Michael~S. Ryoo, Krista Reymann, Kanishka Rao, Karl Pertsch, Igor Mordatch, Henryk Michalewski, Yao Lu, Sergey Levine, Lisa Lee, Tsang{-}Wei~Edward Lee, Isabel Leal, Yuheng Kuang, Dmitry Kalashnikov, Ryan Julian, Nikhil~J. Joshi, Alex Irpan, Brian Ichter, Jasmine Hsu, Alexander Herzog, Karol Hausman, Keerthana Gopalakrishnan, Chuyuan Fu, Pete Florence, Chelsea Finn, Kumar~Avinava Dubey, Danny Driess, Tianli Ding, Krzysztof~Marcin Choromanski, Xi Chen, Yevgen Chebotar, Justice Carbajal, Noah Brown, Anthony Brohan, Montserrat~Gonzalez Arenas, and Kehang Han.
\newblock {RT-2:} vision-language-action models transfer web knowledge to robotic control.
\newblock In \emph{CoRL}, 2023.

\end{thebibliography}
}

\end{document}